\documentclass[10pt,twocolumn,letterpaper]{article}

\usepackage{times}
\usepackage{epsfig}
\usepackage{graphicx}
\usepackage{amsmath}
\usepackage{amssymb}
\usepackage{multirow}
\usepackage[dvipsnames]{xcolor}
\usepackage{authblk}
\usepackage[margin=1in]{geometry}
\usepackage[square,sort,comma,numbers]{natbib}
\definecolor{mygreen}{rgb}{0,0.7,0}
\usepackage{color, soul}

\usepackage{hyperref}

\begin{document}

\date{\vspace{-5ex}}
\title{ELF: Embedded Localisation of Features in pre-trained CNN}

\author[1,2]{Assia Benbihi}
\author[3]{Matthieu Geist}
\author[1,4]{C{\'e}dric Pradalier}
\affil[1]{UMI2958 GeorgiaTech-CNRS}
\affil[2]{Centrale Sup{\'e}lec, Universit{\'e} Paris Saclay, Metz}
\affil[3]{Google Research, Brain Team}
\affil[4]{GeorgiaTech Lorraine}

\maketitle

\begin{abstract}

  This paper introduces a novel feature detector based only on information
  embedded inside a CNN trained on standard tasks (e.g. classification). While
  previous works already show that the features of a trained CNN are suitable
  descriptors, we show here how to extract the feature locations from the
  network to build a detector. This information is computed from the gradient
  of the feature map with respect to the input image. This provides a saliency
  map with local maxima on relevant keypoint locations. Contrary to recent
  CNN-based detectors, this method requires neither supervised training nor
  finetuning. We evaluate how repeatable and how `matchable' the detected
  keypoints are with the repeatability and matching scores. Matchability is
  measured with a simple descriptor introduced for the sake of the
  evaluation. This novel detector reaches similar performances on the standard
  evaluation HPatches dataset, as well as comparable robustness against
  illumination and viewpoint changes on Webcam and photo-tourism images. These
  results show that a CNN trained on a standard task embeds feature location
  information that is as relevant as when the CNN is specifically trained for
  feature detection.
  
\end{abstract}

\section{Introduction}
\begin{figure}[thb]
 \centering
 \includegraphics[width=\linewidth]{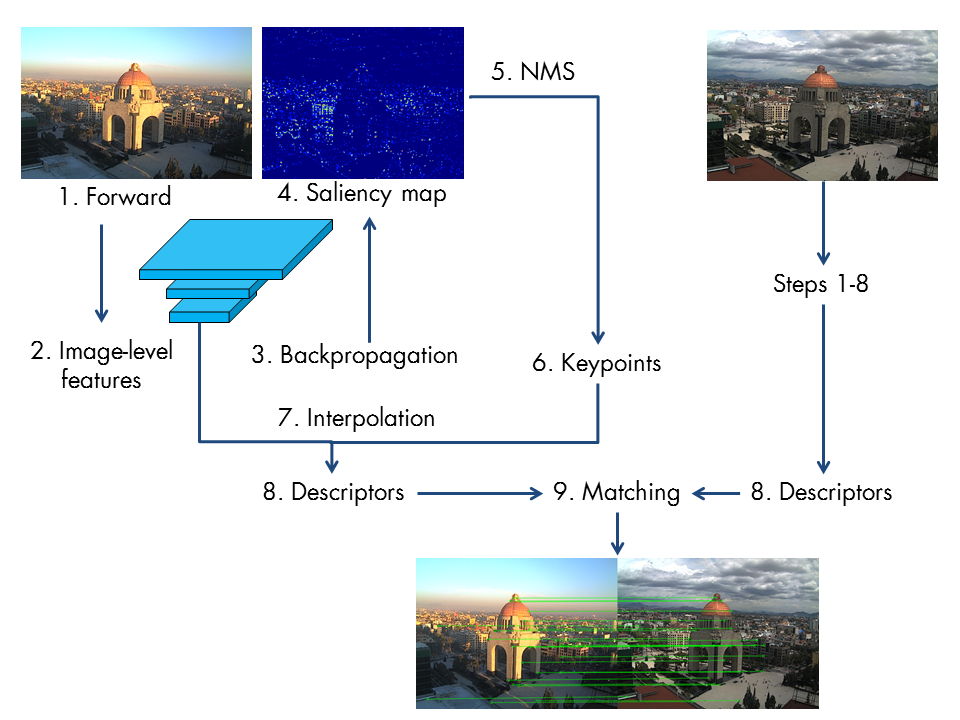}
 \caption{(1-6) Embedded Detector: Given a CNN trained on a standard vision
  task (classification), we backpropagate the feature map back to the
 image space to compute a saliency map. It is thresholded to keep only the
 most informative signal and keypoints are the local maxima. (7-8):
 simple-descriptor.}
 \label{fig:pipeline}
\end{figure}

Feature extraction, description and matching is a recurrent problem in vision
tasks such as Structure from Motion (SfM), visual SLAM, scene recognition and
image retrieval. The extraction consists in detecting image keypoints, then the
matching pairs the nearest keypoints based on their descriptor distance. Even
though hand-crafted solutions, such as SIFT \cite{lowe2004distinctive}, prove
to be successful, recent breakthroughs on local feature detection and
description rely on supervised deep-learning methods \cite{detone18superpoint,
ono2018lf,yi2016lift}. They detect keypoints on saliency maps learned by a
Convolutional Neural Network (CNN), then compute descriptors using another CNN
or a separate branch of it. They all require strong supervision and complex
training procedures: \cite{yi2016lift} requires ground-truth matching keypoints
to initiate the training, \cite{ono2018lf} needs the ground-truth camera pose
and depth maps of the images, \cite{detone18superpoint} circumvents the need
for ground-truth data by using synthetic one but requires a heavy domain
adaptation to transfer the training to realistic images. All these methods
require a significant learning effort. In this paper, we show that a trained
network already embeds enough information to build State-of-the-Art (SoA) detector
and descriptor.

The proposed method for local feature detection needs only a CNN trained on
standard task, such as ImageNet \cite{deng2009imagenet} classification, and no
further training. The detector, dubbed ELF, relies on the features learned by
such a CNN and extract their locations from the feature map gradients. Previous
work already highlights that trained CNN features are relevant descriptors
\cite{fischer2014descriptor} and recent works \cite{balntas2016learning,
han2015matchnet, simo2015discriminative} specifically train CNN to produce
features suitable for keypoint description. However, no existing approach uses a
pre-trained CNN for feature detection. 

\begin{figure}[thb]
 \centering
 \includegraphics[width=\linewidth]{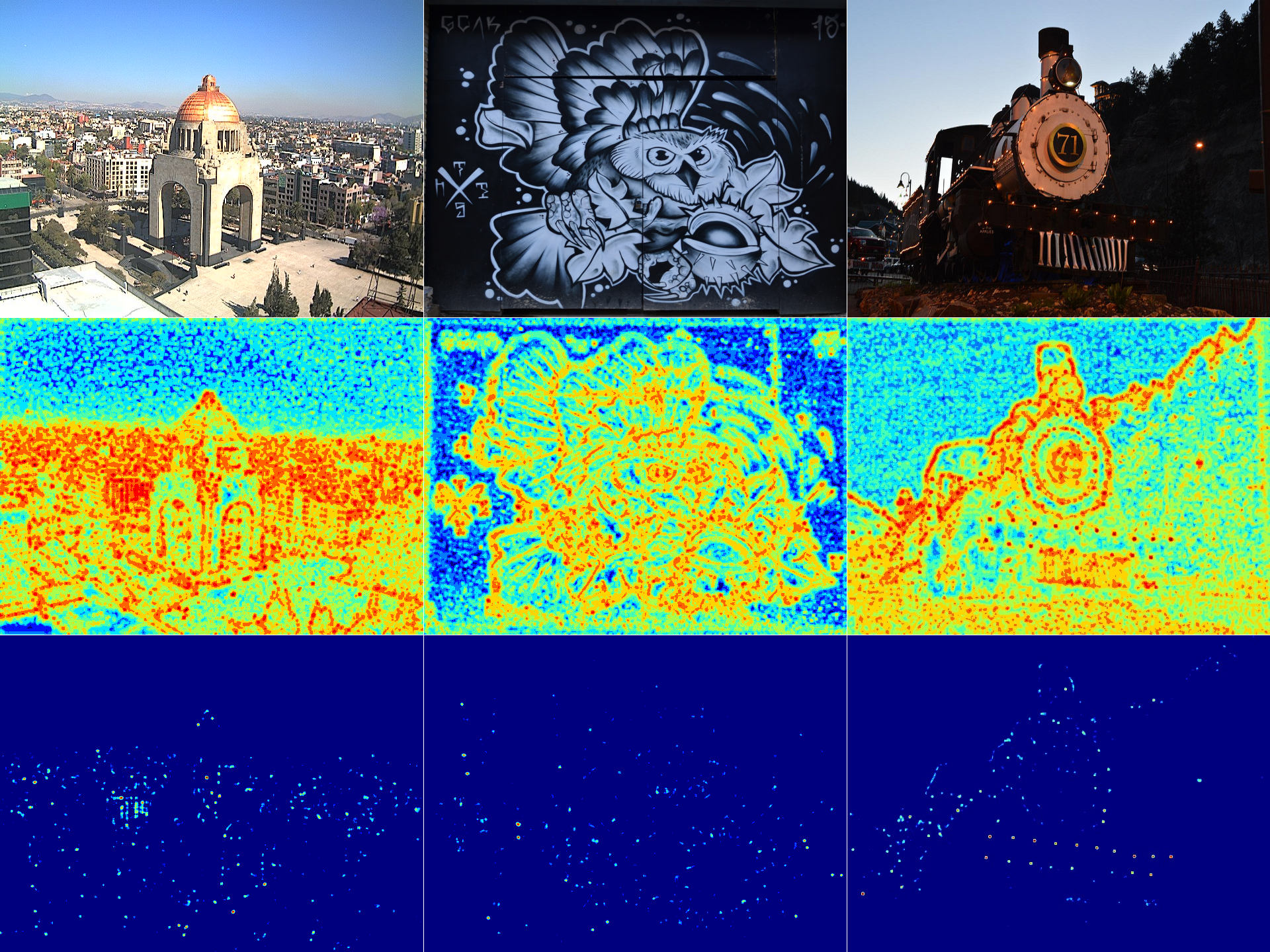}
 \caption{
   Saliency maps thresholding to keep only the most informative location. Top:
   original image. (Left-Right: Webcam \cite{verdie2015tilde}, HPatches
   \cite{balntas2017hpatches}, COCO\cite{lin2014microsoft}) Middle: blurred
   saliency maps. Bottom: saliency map after threshold. (Better seen on a
   computer.)
}
 \label{fig:heatmap_coco}
\end{figure}

ELF computes the gradient of a trained CNN feature map with respect to
\textit{w.r.t} the image: this outputs a saliency map with local maxima on keypoint
positions. Trained detectors learn this saliency map with a CNN whereas we
extract it with gradient computations. This approach is inspired by
\cite{simonyan2013deep} which observes that the gradient of classification
scores \textit{w.r.t} the image is similar to the image saliency map. ELF differs
in that it takes the gradient of feature maps and not the classification score
contrary to existing work exploiting CNN gradients \cite{selvaraju2017grad,
smilkov2017smoothgrad,springenberg2015striving, sundararajan2017axiomatic}. 
These previous works aim at
visualising the learning signal for classification specifically whereas ELF
extracts the feature locations. 
The extracted saliency map is then thresholded to keep only the most relevant
locations and standard Non-Maxima Suppression (NMS) extracts the
final keypoints (Figure \ref{fig:heatmap_coco}). 

ELF relies only on six parameters: 2$\times$2 Gaussian blur
parameters for the automatic threshold level estimation and for the saliency
map denoising; and two parameters for the (NMS) window and the border to
ignore. Detection only requires one forward and one backward passes and takes
$\sim$0.2s per image on a simple Quadro M2200, which makes it suitable for
real-time applications.

ELF is compared to individual detectors with standard \textit{repeatability}
\cite{mikolajczyk2005comparison} but results show that this metric is not
discriminative enough. Most of the existing detectors can extract keypoints
repeated across images with similar repeatability scores. Also, this metric
does not express how `useful' the detected keypoints are: if we sample all
pixels as keypoints, we reach 100\% of \textit{rep.} but the matching may not
be perfect if many areas look alike. Therefore, the detected keypoints
are also evaluated on how `matchable' they are with the \textit{matching score}
\cite{mikolajczyk2005comparison}. This metric requires to describe the
keypoints so we define a simple descriptor: it is based on the interpolation of a CNN
feature map on the detected keypoints, as in \cite{detone18superpoint}. This
avoids biasing the performance by choosing an existing competitive
descriptor. Experiments show that even this simple descriptor reaches competitive
results which comforts the observation of \cite{fischer2014descriptor},
on the relevance of CNN features as
descriptors. More details are provided section 4.1.

ELF is tested on five architectures: three classification networks trained on
ImageNet classification: AlexNet, VGG and Xception
\cite{krizhevsky2012imagenet,simonyan2014very, chollet17xception}, as well as
SuperPoint \cite{detone18superpoint} and LF-Net \cite{ono2018lf} descriptor
networks. Although outside the scope of this paper, this comparison provides
preliminary results of the influence of the network architecture, task and
training data on ELF's performance. Metrics are computed on HPatches
\cite{balntas2017hpatches} for generic performances. We derive two auxiliary
datasets from HPatches to study scale and rotation robustness. Light and 3D
viewpoint robustness analysis are run on the Strecha, Webcam and datasets
\cite{strecha2008benchmarking, verdie2015tilde}. These extensive experiments
show that ELF is on par with other sparse detectors, which suggests that the
feature representation and location information learnt by a CNN to complete a
vision task is as relevant as when the CNN is specifically trained for feature
detection. We additionally test ELF's robustness on 3D reconstruction from
images in the context of the CVPR 2019 Image Matching challenge
\cite{cvpr19challenge}. Once again, ELF is on par with other sparse methods
even though denser methods, e.g. \cite{detone18superpoint}, are more
appropriate for such a task.
Our contributions are the following: 

\begin{itemize} 

  \item We show that a CNN trained on a standard vision task embeds feature
    location in the feature gradients. This information is as
    relevant for feature detection as when a CNN is specifically trained for
    it.

  \item We define a systematic method for local feature detection. Extensive
    experiments show that ELF is on par with other SoA deep trained
    detectors. They also update the previous result from
    \cite{fischer2014descriptor}: self-taught CNN features provide SoA
    descriptors in spite of recent improvements in CNN descriptors 
    \cite{choy2016universal}.

  \item We release the python-based evaluation code to ease future comparison 
    together with ELF code\footnote{ELF code:\url{https://github.com/ELF-det/elf}}.
    The introduced robustness datasets are also made public \footnote{Rotation and scale dataset:
    \url{https://bit.ly/31RAh1S}}.  
\end{itemize}

\section{Related work} 
Early methods rely on hand-crafted detection and
description : SIFT \cite{lowe2004distinctive} detects 3D spatial-scale
keypoints on difference of gaussians and describes them with a 3D Histogram Of
Gradients (HOG). SURF \cite{bay2006surf} uses image integral to speed up the previous
detection and uses a sum of Haar wavelet responses for description. KAZE
\cite{alcantarilla2012kaze} extends the previous multi-scale approach by
detecting features in non-linear scale spaces instead of the classic Gaussian
ones. ORB \cite{rublee2011orb} combines the FAST \cite{rosten2006machine}
detection, the BRIEF \cite{calonder2010brief} description, and improves them to
make the pipeline scale and rotation invariant. MSER-based detector hand-crafts
desired invariance properties for keypoints, and designs a fast algorithm to
detect them \cite{matas2004robust}. Even though these hand-crafted methods have
proven to be successful and to reach state-of-the-art performance for some
applications, recent research focus on learning-based methods.

One of the first learned detector is TILDE \cite{verdie2015tilde}, trained under
drastic changes of light and weather on the Webcam dataset. They use
supervision to learn saliency maps which maxima are keypoint locations.
Ground-truth saliency maps are generated with `good keypoints': they use SIFT
and filter out keypoints that are not repeated in more than 100 images. One
drawback of this method is the need for supervision that relies on another
detector. However, there is no universal explicit definition of what a good
keypoint is. This lack of specification inspires Quad-Networks
\cite{savinov2017quad} to adopt an unsupervised approach: they train a neural
network to rank keypoints according to their robustness to random
hand-crafted transformations. They keep the top/bottom quantile of the
ranking as keypoints. ELF is similar in that it does not requires supervision
but differs in that it does not need to further train the CNN. 

Other learned detectors are trained within full detection/description
pipelines such as LIFT \cite{yi2016lift}, SuperPoint \cite{detone18superpoint}
and LF-Net \cite{ono2018lf}. 
LIFT contribution lies in their original training method of three CNNs.
The detector CNN learns a saliency map where the most salient points are keypoints.
They then crop patches around these keypoints, compute their orientations and 
descriptors with two other CNNs. 
They first train the descriptor with patches around ground-truth matching points 
with contrastive loss, then the orientation CNN together with the descriptor 
and finally with the detector. One drawback of this method is the need for 
ground-truth matching keypoints to initiate the training.
In \cite{detone18superpoint}, the problem is avoided by pre-training 
the detector on a synthetic
geometric dataset made of polygons on which they detect mostly corners. The
detector is then finetuned during the descriptor training on image pairs from COCO
\cite{lin2014microsoft} with synthetic homographies and the correspondence 
contrastive loss introduced in \cite{choy2016universal}.
LF-Net relies on another type of supervision: it uses 
ground-truth camera poses and image depth maps that are easier to compute with 
laser or standard SfM than ground-truth matching keypoints. 
Its training pipeline builds over LIFT and employs the
projective camera model to project detected
keypoints from one image to the other. These keypoint pairs form the ground-truth
matching points to train the network. 
ELF differs in that the CNN model is already trained on a standard task.
It then extracts the relevant information embedded inside
the network for local feature detection, which requires no
training nor supervision.

The detection method of this paper is mainly inspired from the initial
observation in \cite{simonyan2013deep}: given a CNN trained for classification,
the gradient of a class score \textit{w.r.t} the image is the saliency map of
the class object in the input image. A line of works aims at visualizing the
CNN representation by inverting it into the image space through optimization
\cite{mahendran2015understanding,gatys2016image}.
Our work differs in that we backpropagate the feature map itself and not a
feature loss. Following works use these saliency maps to better understand the
CNN training process and justify the CNN outputs. 
Efforts mostly focus on the gradient definitions
\cite{smilkov2017smoothgrad, springenberg2015striving,
sundararajan2017axiomatic, zeiler2014visualizing}.
They differ in the way they handle the backpropagation of the non-linear units
such as Relu. 
Grad-CAM \cite{selvaraju2017grad} introduces a variant where they fuse several
gradients of the classification score \textit{w.r.t} feature maps and not the
image space. Instead, ELF computes the gradient of the feature map, and not a
classification score, \textit{w.r.t} the image. Also we run simple
backpropagation which differs 
in the non-linearity handling: all the signal is backpropagated no matter
whether the feature maps or the gradients are positive or not. Finally, as far
as we know, this is the first work to exploit the localisation information
present in these gradients for feature detection.

The simple descriptor introduced for the sake of the matchability evaluation is
taken from UCN \cite{choy2016universal}. Given a feature map and the keypoints
to describe, it interpolates the feature map on the keypoints location. Using a
trained CNN for feature description is one of the early applications of CNN
\cite{fischer2014descriptor}. Later, research has taken on specifically
training the CNN to generate features suitable for keypoint matching either
with patch-based approaches, among which
\cite{simo2015discriminative,melekhov2016siamese,han2015matchnet,zagoruyko2015learning},
or image-based approaches \cite{taira2018inloc,choy2016universal}. We choose
the description method from UCN~\cite{choy2016universal}, also used by
SuperPoint, for its complexity is only $O(1)$ compared to patch-based
approaches that are $O(N)$ with $N$ the number of keypoints. We favor UCN to
InLoc \cite{taira2018inloc} as it is simpler to compute. The motivation here is
only to get a simple descriptor easy to integrate with all detectors for fair
comparison of the \textit{detector} matching performances. So we overlook
the description performance.

\section{Method}

This section defines ELF, a detection method valid for any trained
CNN. Keypoints are local maxima of a saliency map computed as the feature
gradient \textit{w.r.t} the image. We use the data adaptive Kapur method
\cite{kapur1985new} to automatically threshold the saliency map and keep only the most
salient locations, then run NMS for local maxima detection.

\begin{figure}[thb]
 \centering
 \includegraphics[width=\linewidth]{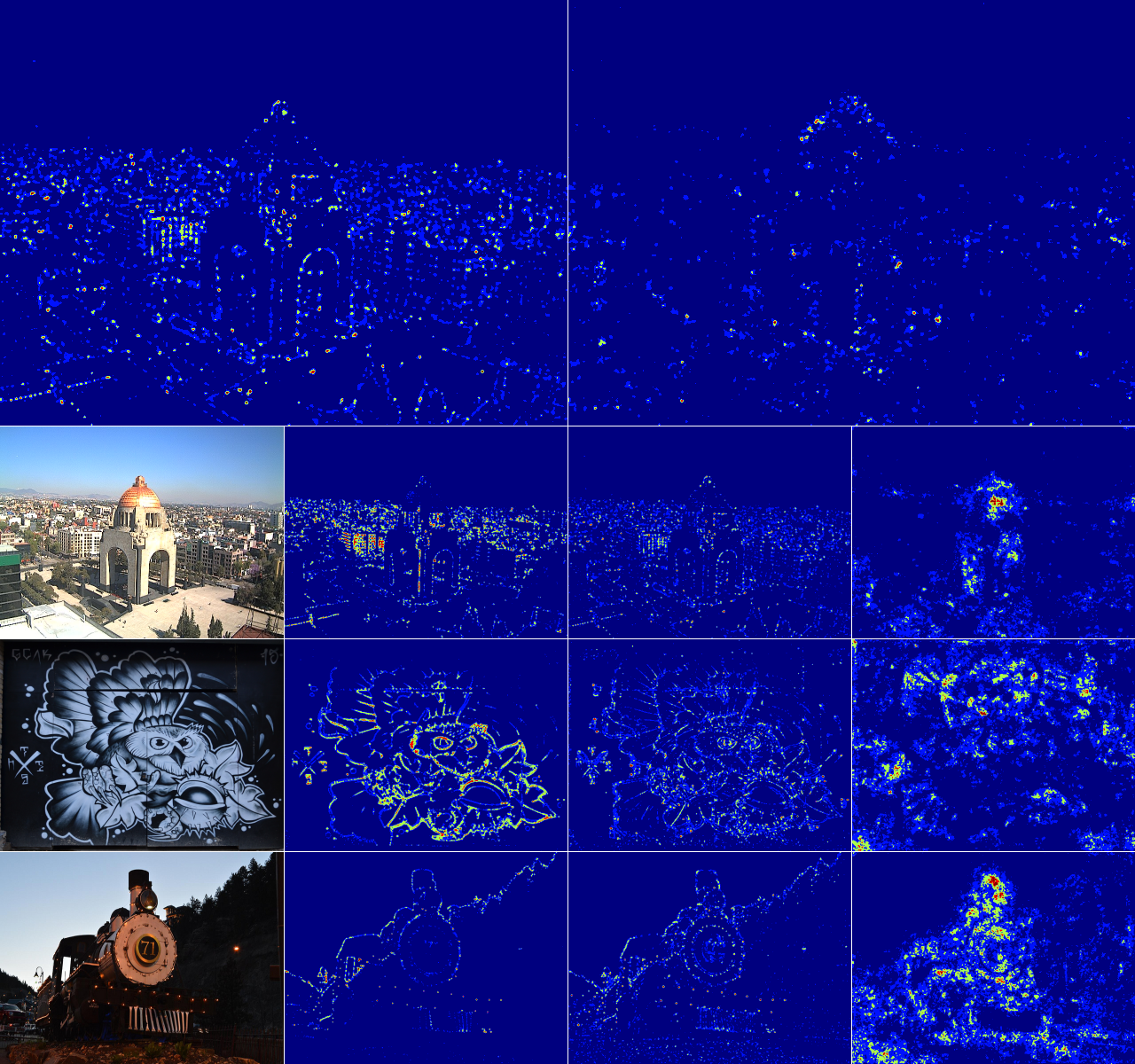}
  \caption{(Bigger version Figure \ref{fig:big_saliency_coco}.)
  Saliency maps computed from the feature map gradient 
 $\left| ^TF^l(x) \cdot \frac{\partial F^l}{\partial \mathbf{I}} \right|$.
 Enhanced image contrast for better visualisation.
 Top row: gradients of VGG $pool_2$ and $pool_3$ show a loss of resolution 
 from $pool_2$ to $pool_3$.
 Bottom: $(pool_i)_{i \in [1,2,5]}$ of VGG on Webcam, HPatches and Coco images.
 Low level saliency maps activate accurately whereas higher saliency maps are blurred.}
 \label{fig:saliency_coco}
\end{figure}

\subsection{Feature Specific Saliency}

We generate a saliency map that activates on the
most informative image region for a specific CNN feature level $l$.
Let $\mathbf{I}$ be a vector image of dimension $D_I = H_I \cdot W_I \cdot C_I$.
Let $F^l$ be a vectorized feature map of dimension $D_F= H_l \cdot W_l \cdot C_l$.
The saliency map $S^l$, of dimension $D_I$, is
$S^l(\mathbf{I})=\left| ^tF^l(\mathbf{I}) \cdot \nabla_I F^l \right|$,
with $\nabla_I F^l$ a $D_F \times D_I$ matrix.

The saliency activates on the image regions that contribute the most to the
feature representation $F^l(\mathbf{I})$. The term $\nabla_I F^l$ explicits the
correlation between the feature space of $F^l$ and the image space in general.
The multiplication by $F^l(\mathbf{I})$ applies the correlation to the features
$F^l(\mathbf{I})$ specifically and generate a visualisation in image space
$S^l(\mathbf{I})$. From a geometrical point of view, this operation can be seen
as the projection $\nabla_I F^l$ of a feature
signal $F^l(\mathbf{I})$ into the image space. From a signal processing
approach, $F^l(\mathbf{I})$ is an input signal filtered through $\nabla_I F^l$
into the image space. 
If $C_I>1$, $S^l$ is converted into a grayscale
image by averaging it across channels.

\subsection{Feature Map Selection}

We provide visual guidelines to choose the feature level $l$ so that $F^l$
still holds high resolution localisation information while providing a useful
high-level representation. 

CNN operations such as convolution and pooling increase the receptive field of
feature maps while reducing their spatial dimensions. 
This means that $F^{l}$ has less spatial resolution than $F^{l-1}$ and the backpropagated 
signal $S^l$ ends up more spread than $S^{l-1}$. This is similar to 
when an image is too enlarged and it can be observed in Figure
\ref{fig:saliency_coco}, which shows the gradients of the VGG feature maps.
On the top row, $pool_2$'s gradient (left) better captures
the location details of the dome whereas $pool_3$'s gradient (right) is more spread. 
On the bottom rows, the images lose their resolution as we go higher in the network. 
Another consequence of this resolution loss is that small features are not 
embedded in $F^l$ if $l$ is too high. This would reduce the space of potential
keypoint to only large features which would hinder the method.
This observation motivates us to favor low-level feature maps for feature
detection. We chose the final $F^l$ by taking the highest $l$ which provides
accurate localisation. This is visually observable by sparse high intensity
signal contrary to the blurry aspect of higher layers. 

\subsection{Automatic Data-Adaptive Thresholding}

The threshold is automatic and adapts to the saliency map distribution to keep
only the most informative regions. Figure \ref{fig:heatmap_coco} shows saliency
maps before and after thresholding using Kapur's method \cite{kapur1985new},
which we briefly recall below. It chooses the threshold to maximize the
information between the image background and foreground \textit{i.e.}
the pixel distribution below and above the threshold. This method is especially
relevant in this case as it aims at maintaining as much information on the
distribution above the threshold as possible. This distribution describes the
set of local maxima among which we choose our keypoints. More formally, for an
image $\mathbf{I}$ of $N$ pixels with $n$ sorted gray levels and $(f_i)_{i \in
n}$ the corresponding histogram, $p_i=\frac{f_i}{N}$ is the empirical
probability of a pixel to hold the value $f_i$. Let $s \in n$ be a threshold
level and $A,B$ the empirical background and foreground distributions. The
level $s$ is chosen to maximize the information between A and B and the
threshold value is set to $f_s$: $A = \left(
\frac{p_i}{\sum_{i<s}pi}\right)_{i<s}$ and $B =
\left(\frac{p_i}{\sum_{i>=s}pi}\right)_{i>s}$. For better results, we blur the
image with a Gaussian of parameters $(\mu_{thr}, \sigma_{thr})$ before
computing the threshold level.

Once the threshold is set, we denoise the image with a second Gaussian blur of
parameters $(\mu_{noise}, \sigma_{noise})$ and run standard NMS (the same as
for SuperPoint) where we iteratively select decreasing global maxima while ensuring
that their nearest neighbor distance is higher than the window $w_{\textrm{NMS}} \in \mathbb{N}$. 
Also we ignore the $b_{\textrm{NMS}} \in \mathbb{N}$ pixels around the image
border.

\subsection{Simple descriptor}

As mentioned in the introduction, the repeatability score does not discriminate
among detectors anymore. So they are also evaluated on how `matchable' their
detected keypoints are with the matching score. To do so, the ELF detector is
completed with a simple descriptor inspired by SuperPoint's descriptor. The use
of this simple descriptor over existing competitive ones avoids unfairly
boosting ELF's perfomance. Inspired by SuperPoint, we interpolate a CNN feature
map on the detected keypoints. Although simple, experiments show that this
simple descriptor completes ELF into a competitive feature
detection/description method. 

The feature map used for description may be different from the one for
detection. High-level feature maps have wider receptive field hence take
higher context into account for the description of a pixel location. This leads
to more informative descriptors which motivates us to favor higher level maps.
However we are also constrained by the loss of resolution previously described:
if the feature map level is too high, the interpolation of the descriptors generate
vector too similar to each other. For example, the VGG $pool_4$ layer produces
more discriminative descriptors than $pool_5$ even though $pool_5$ embeds
information more relevant for classification. Empirically we observe that there
exists a layer level $l'$ above which the description performance stops
increasing before decreasing. This is measured through the matching score
metric introduced in \cite{mikolajczyk2005comparison}. The final choice of the
feature map is done by testing some layers $l'>l$ and select the lowest feature
map before the descriptor performance stagnates. 

The compared detectors are evaluated with both
their original descriptor and this simple one. We detail the motivation behind
this choice: detectors may be biased to sample keypoints that their respective
descriptor can describe `well' \cite{yi2016lift}. So it is fair to compute the matching score
with the original detector/descriptor pairs. However, a detector can sample
`useless points' (e.g. sky pixels for 3d reconstructions) that its descriptor
can characterise `well'. In this case, the descriptor `hides' the detector
default. This motivates the integration of a common independent descriptor with
all detectors to evaluate them. Both approaches are run since each is as fair
as the other.

\section{Experiments}

This section describes the evaluation metrics and datasets as well as the
method's tuning.
Our method is compared to detectors with available public code: the fully hand-crafted
SIFT \cite{lowe2004distinctive}, SURF \cite{bay2006surf}, ORB
\cite{rublee2011orb}, KAZE \cite{alcantarilla2012kaze}, the
learning-based LIFT \cite{yi2016lift}, SuperPoint \cite{detone18superpoint},
LF-Net \cite{ono2018lf}, the individual detectors TILDE \cite{verdie2015tilde},
MSER \cite{matas2004robust}.

\subsection{Metrics}

We follow the standard validation guidelines \cite{mikolajczyk2005comparison}
that evaluates the detection performance with \textit{repeatability (rep)}. It
measures the percentage of keypoints common to both images. We also compute 
the \textit{matching
score (ms)} as an additional \textit{detector} metric. It captures the
percentage of keypoint pairs that are nearest neighbours in both image space and
descriptor space i.e. the ratio of keypoints correctly matched. For fair
completeness, the mathematical definitions of the metrics are provided in
Appendix and their implementation in the soon-to-be released code.

A way to reach perfect
\textit{rep} is to sample all the pixels or sample them with a
frequency higher than the distance threshold $\epsilon_{kp}$ of the metric. One
way to prevent the first flaw is to limit the number of keypoints but it does
not counter the second. Since detectors are always used together with
descriptors, another way to think the detector evaluation is: \textit{'a good
keypoint is one that can be discriminatively described and matched'}. One could
think that such a metric can be corrupted by the descriptor. But
we ensure that a detector flaw cannot be hidden by a very performing descriptor
with two guidelines. One experiment must evaluate all detector with one fixed
descriptor (the simple one defined in 3.4). Second, \textit{ms} can never be higher
than \textit{rep} so a detector with a poor \textit{rep} leads to a poor
\textit{ms}.

Here the number of detected keypoints is limited to 500 for all methods. As
done in \cite{detone18superpoint,ono2018lf}, we replace the overlap score in
\cite{mikolajczyk2005comparison} to compute correspondences with the 5-pixel
distance threshold. Following \cite{yi2016lift}, we also modify the matching
score definition of \cite{mikolajczyk2005comparison} to run a greedy
bipartite-graph matching on all descriptors and not just the descriptor pairs
for which the distance is below an arbitrary threshold. We do so to be able to
compare all state-of-the-art methods even when their descriptor dimension and
range vary significantly. (More details in Appendix.)

\subsection{Datasets}
All images are resized to the 480$\times$640 pixels and the image
pair transformations are rectified accordingly.

\begin{figure}[thb]
 \centering
 \includegraphics[width=\linewidth]{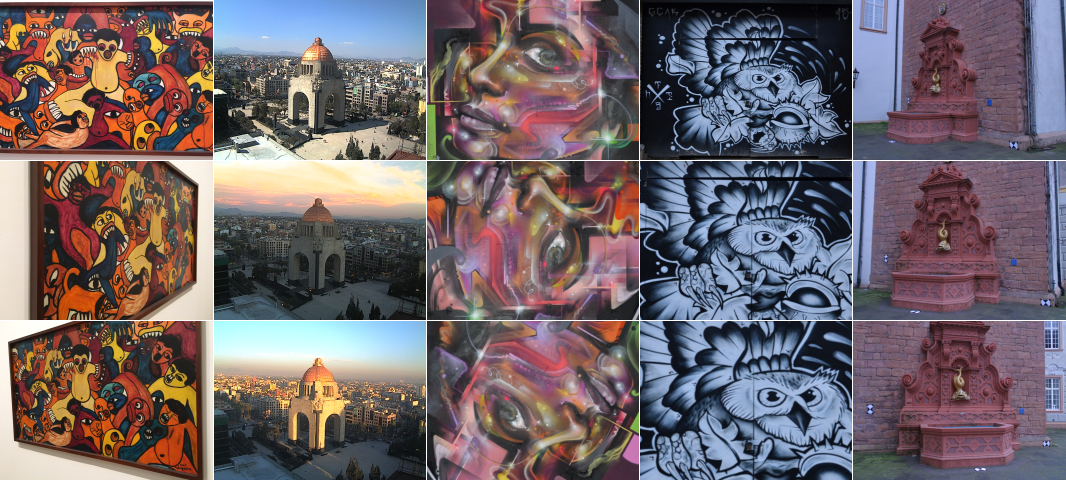}
 \caption{Left-Right: HPatches: planar viewpoint. 
 Webcam: light.
 HPatches: rotation. HPatches: scale. Strecha: 3D viewpoint.} 
 \label{fig:datasets}
\end{figure}

\textbf{General performances.}
The HPatches dataset \cite{balntas2017hpatches} gathers a subset of standard
evaluation images such as DTU and OxfordAffine
\cite{aanaes2012interesting,mikolajczyk2005performance}: it provides a total of
696 images, 6 images for 116 scenes and the corresponding homographies between
the images of a same scene. For 57 of these scenes, the main changes are
photogrammetric and the remaining 59 show significant geometric deformations
due to viewpoint changes on planar scenes.

\textbf{Illumination Robustness.}
The Webcam dataset \cite{verdie2015tilde}
gathers static outdoor scenes with drastic natural light changes contrary to
HPatches which mostly holds artificial light changes in indoor scenes. 

\textbf{Rotation and Scale Robustness.}
We derive two datasets from HPatches. For each of the 116 scenes,
we keep the first image and rotate it with angles from $0^{\circ}$ to
$210^{\circ}$ with an interval of $40^{\circ}$. Four zoomed-in version of the
image are generated with scales $[1.25, 1.5, 1.75, 2]$. We release these two
datasets together with their ground truth homographies for future comparisons.

\textbf{3D Viewpoint Robustness.}
We use three Strecha scenes \cite{strecha2008benchmarking} with increasing
viewpoint changes: \textit{Fountain, Castle entry, Herzjesu-P8}.  The viewpoint
changes proposed by HPatches are limited to planar scenes which does not
reflect the complexity of 3D structures. Since the ground-truth depths are not
available anymore, we use COLMAP \cite{schonberger2016structure} 3D
reconstruction to obtain ground-truth scaleless depth. We release the obtained
depth maps and camera poses together with the evaluation code. 
ELF robustness is additionally tested in the CVPR19 Image Matching Challenge
\cite{cvpr19challenge} (see results sections).

\subsection{Baselines}

We describe the rationale behind the evaluation.
The tests run on a
QuadroM2200 with Tensorflow 1.4, Cuda8, Cudnn6 and Opencv3.4.
We use the OpenCV implementation of SIFT, SURF, ORB, KAZE, MSER with
the default parameters and the author's code for TILDE, LIFT, 
SuperPoint, LF-Net with the provided models and parameters. 
When comparing detectors in the feature matching
pipeline, we measure their matching score with both their original descriptor and
ELF simple descriptor. For MSER and TILDE, we use the
VGG simple descriptor.

\textbf{Architecture influence.} ELF is tested on five networks: three
classification ones trained on ImageNet (AlexNet, VGG, Xception
\cite{krizhevsky2012imagenet, simonyan2014very,chollet17xception}) as well as
the trained SuperPoint's and LF-Net's descriptor ones. We call each variant
with the network's names prefixed with ELF as in saliency. The paper compares
the influence of i) architecture for a fixed task (ELF-AlexNet
\cite{krizhevsky2012imagenet} \textit{vs.} ELF-VGG \cite{simonyan2014very}
\textit{v.s.} ELF-Xception \cite{chollet17xception}), ii) the task (ELF-VGG
\textit{vs.} ELF-SuperPoint (SP) descriptor), iii) the training dataset
(ELF-LFNet on phototourism \textit{vs.} ELF-SP on MS-COCO). This study is
being refined with more independent comparisons of tasks, datasets and
architectures soon available in a journal extension.

We use the author's code and pre-trained models which we
convert to Tensorflow \cite{abadi2016tensorflow} except for LF-Net.
We search the blurring parameters 
$(\mu_{thr}, \sigma_{thr})$, $(\mu_{noise}, \sigma_{noise})$
in the range $ [\![3,21]\!]^2$ and the NMS parameters $(w_{NMS},
b_{NMS})$ in $[\![4,13]\!]^2$. 

\textbf{Individual components comparison.} Individual detectors are compared 
with the matchability of their detection and the description of the
simple VGG-pool3 descriptor. This way, the \textit{m.s.} only depends on the
detection performance since the description is fixed for all detectors.
The comparison between ELF and recent deep methods raises the question of
whether triplet-like losses are relevant to train CNN descriptors. Indeed,
these losses constrain the CNN features directly so that matching keypoints are
near each other in descriptor space. Simpler loss, such as cross-entropy for
classification, only the constrain the CNN output on the task while leaving the
representation up to the CNN.

ELF-VGG detector is also integrated with existing descriptors. This evaluates 
how useful the CNN self-learned feature localisation compares with the
hand-crafted and the learned ones. 

\textbf{Gradient Baseline.}
Visually, the feature gradient map is reminiscent of the image gradients computed
with the Sobel or Laplacian operators. We run two variants of our pipeline 
where we replace the feature gradient with them.
This aims at showing whether CNN feature gradients embed more information than
image intensity gradients.

\section{Results}

Experiments show that ELF compares with the state-of-the-art on HPatches and
demonstrates similar robustness properties with recent learned methods.
It generates saliency maps visually akin to a Laplacian on very structured
images (HPatches) but proves to be more robust on outdoor scenes with natural
conditions (Webcam). When integrated with existing feature descriptors, ELF
boosts the matching score. Even integrating ELF
simple descriptor improves it with the exception of SuperPoint for which
results are equivalent. This sheds new light on the representations learnt by
CNNs and suggests that deep description methods may underexploit the
information embedded in their trained networks. Another suggestion may be that
the current metrics are not relevant anymore for deep learning methods. Indeed,
all can detect repeatable keypoints with more or less the same performances.
Even though the matchability of the points (\textit{m.s}) is a bit more
discriminative, neither express how `useful' the \textit{kp} are for the
end-goal task. One way to do so is to evaluate an end-goal task (\textit{e.g.}
Structure-from-Motion). However, for the evaluation to be rigorous all the
other steps should be fixed for all papers. Recently, the Image Matching CVPR19
workshop proposed such an evaluation but is not fully automatic yet. 
These results also challenge whether current descriptor-training loss are a strong
enough signal to constrain CNN features better than a simple cross-entropy.

\begin{figure}[htb]
 \centering
 \hbox{ \includegraphics[width=\linewidth]{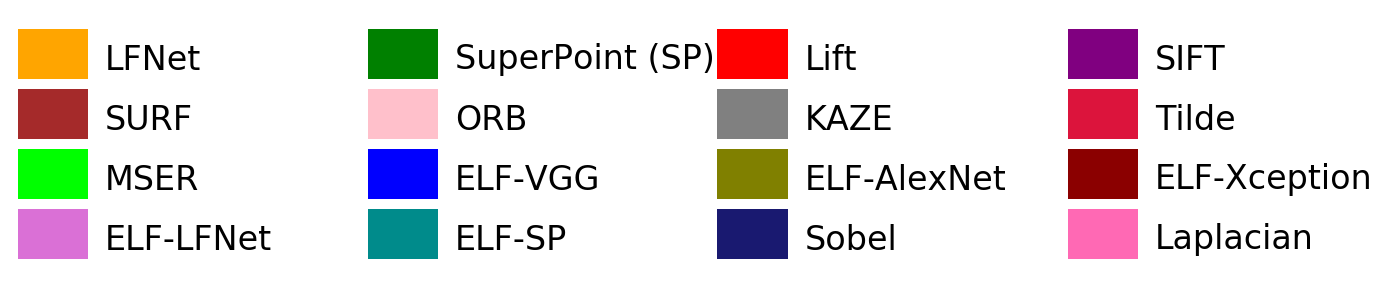}}
 \hbox{ \includegraphics[width=\linewidth]{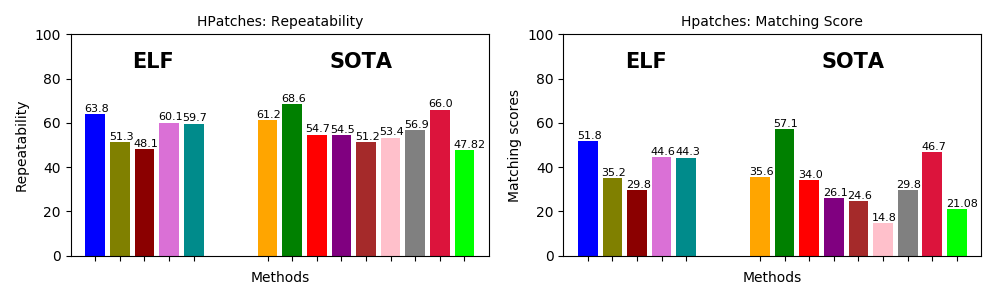}}
 \hbox{ \includegraphics[width=\linewidth]{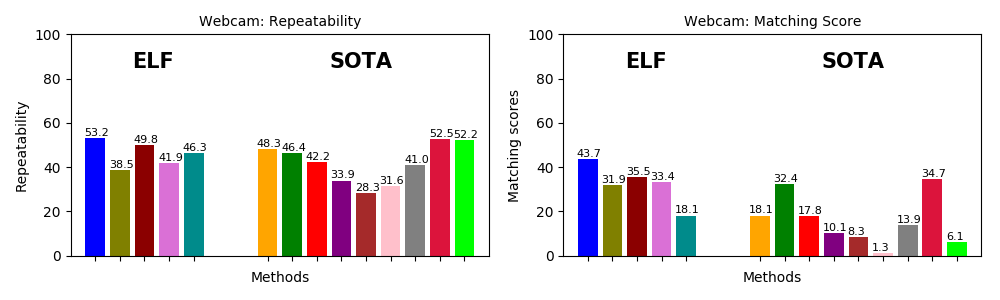}}
 \caption{Top-Down: HPatches-Webcam. Left-Right: repeatability, matching score.}
\label{fig:hpatch_gle_perf}
\end{figure}

The tabular version of the following results is provided in Appendix. The graph
results are better seen with color on a computer screen. Unless
mentioned otherwise, we compute repeatability for each detector, and the
matching score of detectors with their respective descriptors, when they have
one. We use ELF-VGG-$pool_4$ descriptor for TILDE, MSER, ELF-VGG,
ELF-SuperPoint, and ELF-LFNet. We use AlexNet and Xception feature
maps to build their respective simple descriptors. The meta-parameters for each
variants are provided in Appendix. 

\textbf{General performances.} Figure \ref{fig:hpatch_gle_perf} (top) shows
that the \textit{rep} variance is low across detectors whereas \textit{ms} is
more discriminative, hence the validation method (Section 4.1). On HPatches,
SuperPoint (SP) reaches the best \textit{rep}-\textit{ms} [68.6, 57.1] closely
followed by ELF (e.g. ELF-VGG: [63.8, 51.8]) and TILDE [66.0, 46.7]. In
general, we observe that learning-based methods all outperform hand-crafted
ones. Still, LF-Net and LIFT curiously underperform on HPatches: one reason may
be that the data they are trained on differs too much from this one. LIFT is
trained on outdoor images only and LF-Net on either indoor or outdoor datasets,
whereas HPatches is made of a mix of them. We compute metrics for both LF-Net
models and report the highest one (indoor). Even though LF-Net and LIFT fall
behind the top learned methods, they still outperform hand-crafted ones
which suggests that their framework learn feature specific information
that hand-crafted methods can not capture. This supports the recent direction
towards trained detectors and descriptors.

\textbf{Light Robustness}
Again, \textit{ms} is a better discriminant on Webcam than \textit{rep}
(Figure \ref{fig:hpatch_gle_perf} bottom). ELF-VGG reaches
top \textit{rep}-\textit{ms} [53.2, 43.7] closely followed by TILDE [52.5,
34.7] which was the state-of-the-art detector. 

Overall, there is a performance degradation ($\sim$20\%) from HPatches to
Webcam. HPatches holds images with standard features such as corners that
state-of-the-art methods are made to recognise either by definition or by
supervision. There are less such features in the Webcam dataset because of the
natural lighting that blurs them. Also there are strong intensity variations
that these models do not handle well. One reason may be that the learning-based
methods never saw such lighting variations in their training set. But this
assumption is rejected as we observe that even SuperPoint, which is trained on
Coco images, outperforms LIFT and LF-Net, which are trained on outdoor images.
Another justification can be that what matters the most is the pixel
distribution the network is trained on, rather than the image content. The top
methods are classifier-based ELF and SuperPoint: the first ones are trained
on the huge Imagenet dataset and benefit from heavy data augmentation.
SuperPoint also employs a considerable data strategy to train their network.
Thus these networks may cover a much wider pixel distribution which would
explain their robustness to pixel distribution changes such as light
modifications. 

\textbf{Architecture influence}
ELF is tested on three classification networks as well as
the descriptor networks of SuperPoint and LF-Net (Figure
\ref{fig:hpatch_gle_perf}, bars under `ELF').

For a fixed training task (classification) on a fixed dataset (ImageNet),
VGG, AlexNet and Xception are compared. As could be expected, the
network architecture has a critical impact on the detection and ELF-VGG
outperforms the other variants. The \textit{rep} gap can be explained by the fact that
AlexNet is made of wider convolutions than VGG, which induces a higher loss of
resolution when computing the gradient. As for \textit{ms}, the higher
representation space of VGG may help building more informative features which
are a stronger signal to backpropagate. This could also justify why ELF-VGG
outperforms ELF-Xception that has less parameters. Another explanation
is that ELF-Xception's gradient maps seem smoother. Salient
locations are then less emphasized which makes the keypoint detection harder.
One could hint at the depth-wise convolution to explain this visual aspect but
we could not find an experimental way to verify it.
Surprisingly, ELF-LFNet outperforms the original LF-Net on both HPatches
and Webcam and ELF-SuperPoint variant reaches similar results as the original.

\begin{figure}[thb]
 \centering
 \includegraphics[width=\linewidth]{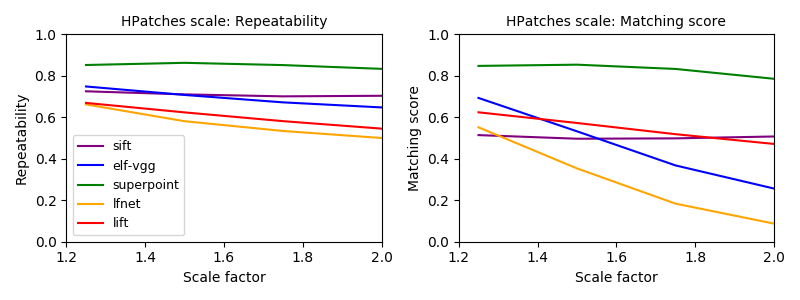}
 \caption{HPatches scale. Left-Right: rep, ms.}
 \label{fig:robust_scale}
\end{figure}

\textbf{Scale Robustness.} ELF-VGG is compared with state-of-the art detectors and
their respective descriptors (Figure \ref{fig:robust_scale}). Repeatability is
mostly stable for all methods: SIFT and SuperPoint are the
most invariant whereas ELF follows the same variations as LIFT and
LF-Net. Once again, \textit{ms} better assesses the detectors performance:
SuperPoint is the most robust to scale changes, followed by LIFT and SIFT.
ELF and LF-Net lose 50\% of their matching score with the
increasing scale. It is surprising to observe that LIFT is more scale-robust
than LF-Net when the latter's global performance is higher. 
A reasonable explanation is that LIFT detects keypoints at 21 scales of the
same image whereas LF-Net only runs its detector CNN on 5 scales. Nonetheless,
ELF outperforms LF-Net without manual multi-scale processing.

\begin{figure}[thb]
 \centering
 \includegraphics[width=\linewidth]{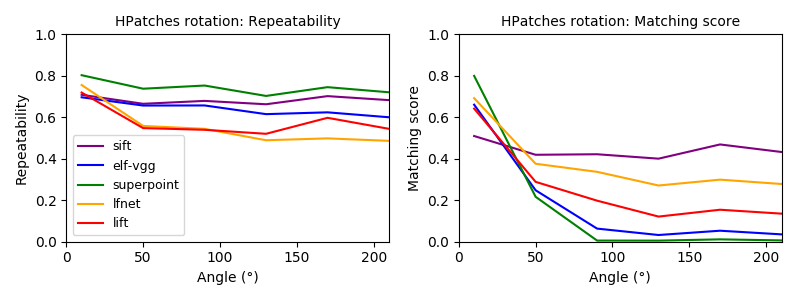}
 \caption{HPatches rotation. Left-Right: rep, ms.}
 \label{fig:robust_rotation}
\end{figure}

\textbf{Rotation Robustness.} Even though \textit{rep} shows little variations
(Figure \ref{fig:robust_rotation}), all learned methods' \textit{ms} crash
while only SIFT survives the rotation changes. This can be explained by the
explicit rotation estimation step of SIFT. However LIFT and LF-Net also run
such a computation. This suggests that either SIFT's hand-crafted
orientation estimation is more accurate or that 
HOG are more rotation invariant than learned
features. LF-Net still performs better than LIFT: this may be because it learns
the keypoint orientation on the keypoint features representation rather than
the keypoint pixels as done in LIFT. Not surprisingly, ELF simple descriptor is
not rotation invariant as the convolutions that make the CNN are not. This also
explains why SuperPoint also crashes in a similar manner. These results suggest
that the orientation learning step in LIFT and LF-Net is needed but its
robustness could be improved. 

\begin{figure}[thb]
 \centering
 \includegraphics[width=\linewidth]{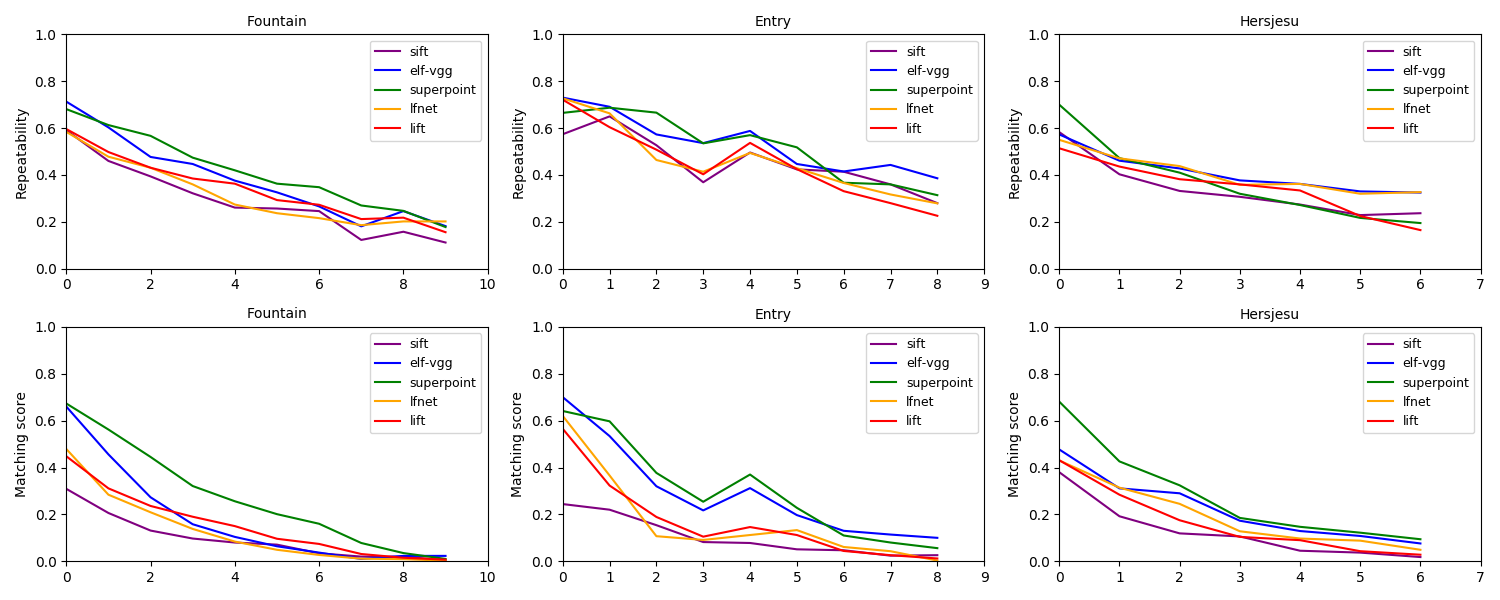}
 \caption{Robustness analysis: 3D viewpoint.}
 \label{fig:robust_strecha}
\end{figure}
\textbf{3D Viewpoint Robustness.} While SIFT shows a clear advantage of
pure-rotation robustness, it displays similar degradation as other methods on
realistic rotation-and-translation on 3D structures. Figure
\ref{fig:robust_strecha} shows that all methods degrade uniformly. One could
assume that this small data sample is not representative enough to run such
robustness analysis. However, we think that these results rather suggest that
all methods have the same robustness to 3D viewpoint changes. Even though
previous analyses allows to rank the different feature matching pipelines, each
has advantages over others on certain situations: ELF or SuperPoint on
general homography matches, or SIFT on rotation robustness. This is why this paper 
only aims at showing ELF reaches the same performances and shares similar
properties to existing methods as there is no generic ranking criteria.
The recent evaluation run by the CVPR19 Image Matching Challenge
\cite{cvpr19challenge} supports the previous conclusions.

\begin{figure}[thb]
 \centering
 \includegraphics[width=\linewidth]{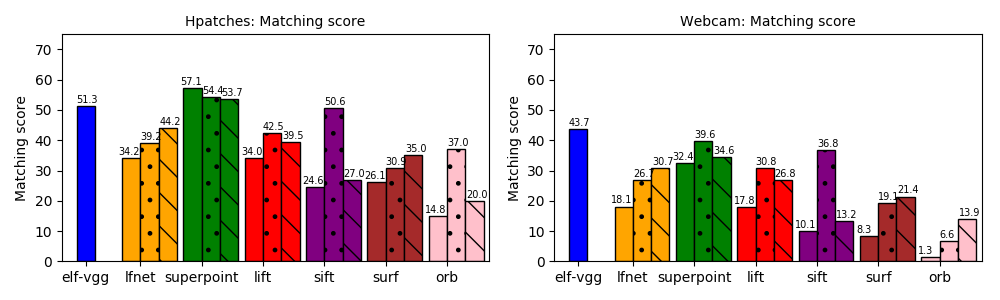}
 \caption{Left-Middle-Right bars: original method, integration of ELF detection, 
  integration of ELF description.}
 \label{fig:ind_component}
\end{figure}

\textbf{Individual components performance.} First, all methods' descriptor
are replaced with the simple ELF-VGG-$pool_3$ one. We then compute
their new \textit{ms} and compare it to ELF-VGG on HPatches and Webcam (Figure
\ref{fig:ind_component}, stripes). The description is based on $pool_3$ instead
of $pool_4$ here for it produces better results for the other methods while
preserving ours. ELF reaches higher \textit{ms} [51.3] for all methods except
for SuperPoint [53.7] for which it is comparable. This shows that ELF is as
relevant, if not more, than previous hand-crafted or learned detectors. This
naturally leads to the question: \textit{'What kind of keypoints does ELF detect
?'} There is currently no answer to this question as it is complex to
explicitly characterize properties of the pixel areas around keypoints. Hence
the open question \textit{'What makes a good keypoint ?'} mentioned at the
beginning of the paper. Still, we observe that ELF activates mostly on high
intensity gradient areas although not all of them. One explanation is that as
the CNN is trained on the vision task, it learns to ignore image regions
useless for the task. This results in killing the gradient
signals in areas that may be unsuited for matching. 

Another surprising observation regards CNN descriptors: SuperPoint (SP)
keypoints are described with the SP descriptor in one hand and the simple ELF-VGG one
in the other hand. Comparing the two resulting matching scores is one way to
compare the SP and ELF descriptors. Results show that both approaches lead to
similar \textit{ms}. This result is surprising because SP specifically trains a
description CNN so that its feature map is suitable for keypoint description
\cite{choy2016universal}. In VGG training, there is no explicit constraints on
the features from the cross-entropy loss. Still, both feature maps reach
similar numerical description performance. This raises the question of whether
contrastive-like losses, which input are CNN features, can better constrain the
CNN representation than simpler losses, such as cross-entropy, which inputs are
classification logits.
This also shows that there is more to CNNs than only the task they are trained
on: they embed information that can prove useful for unrelated tasks.
Although the simple descriptor was defined for evaluation purposes, these
results demonstrate that it can be used as a description baseline for feature
extraction. 

The integration of ELF detection with other methods'
descriptor (Figure \ref{fig:ind_component}, circle) boosts 
the \textit{ms}. \cite{yi2016lift}~previously suggested that there may be a
correlation between the detector and the descriptor within a same method, i.e.
the LIFT descriptor is trained to describe only the keypoints output by its
detector. However, these results show that ELF can easily be
integrated into existing pipelines and even boost their performances.

\begin{figure}[htb]
 \centering
 \hbox{ \includegraphics[width=\linewidth]{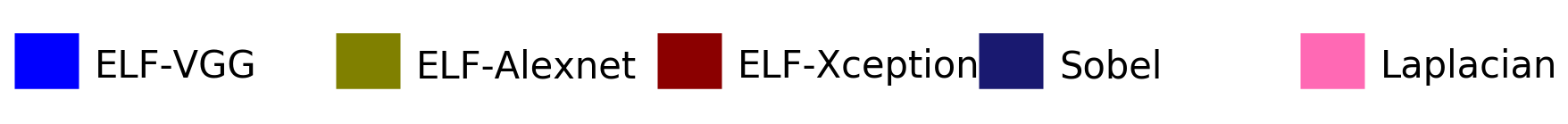}}
 \hbox{ \includegraphics[width=\linewidth]{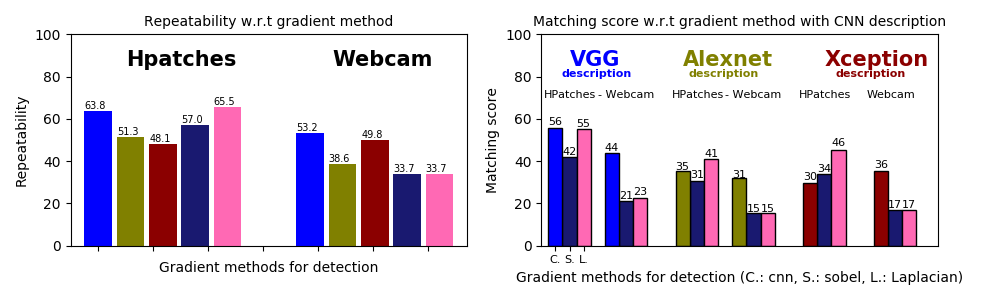}}
 \caption{Gradient baseline.}
 \label{fig:gradient_perf}
\end{figure}

\begin{figure}[thb]
 \centering
 \includegraphics[height=3cm]{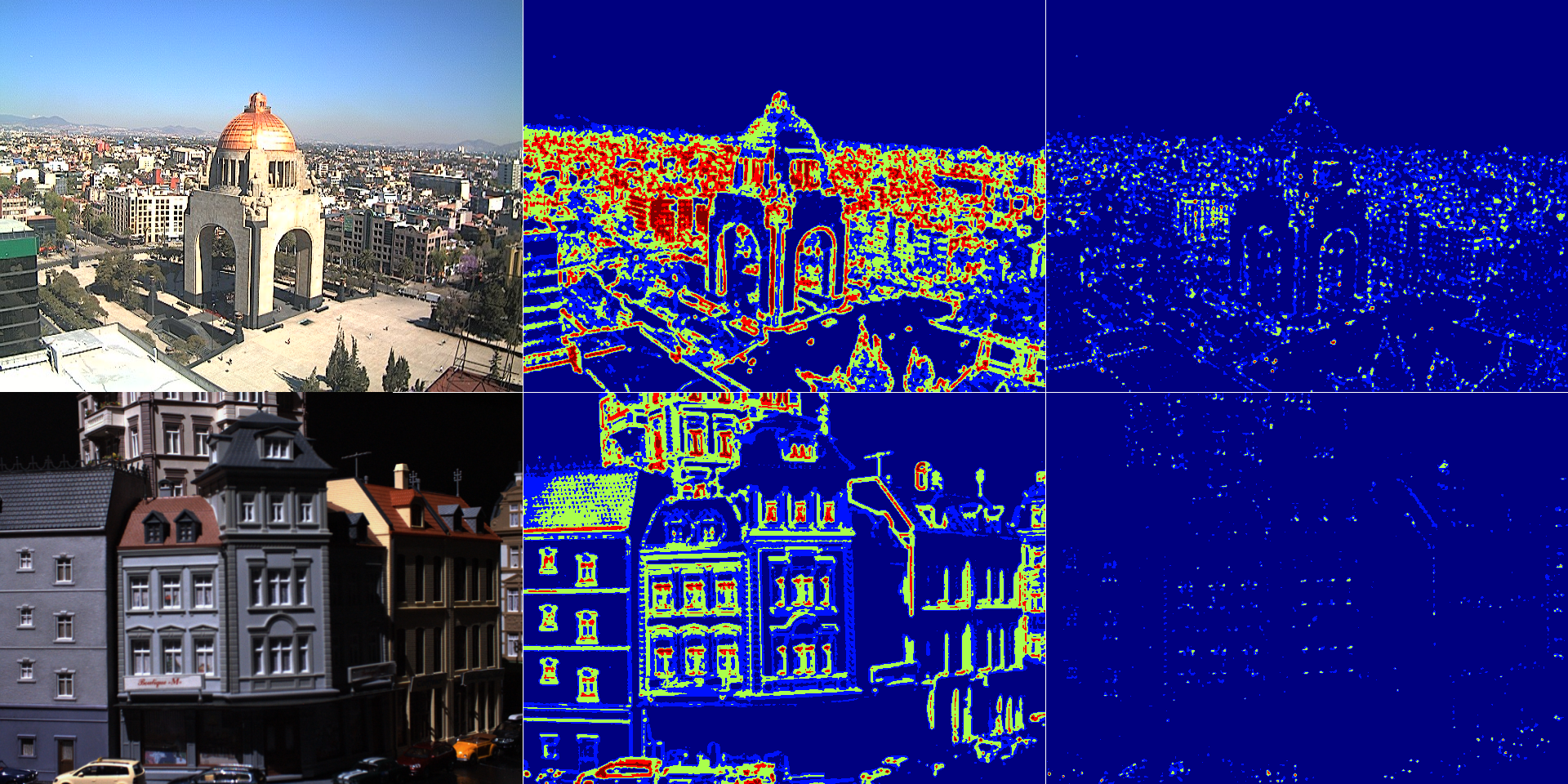}
  \caption{Feature gradient (right) provides a sparser signal than Laplacian
  (middle) which is more selective of salient areas.}
 \label{fig:sobel_visu}
\end{figure}

\textbf{Gradient Baseline} The saliency map used in ELF is replaced 
with simple Sobel or
Laplacian gradient maps. The rest of the detection pipeline stays the same 
and we compute their performance (Figure
\ref{fig:gradient_perf} Left). They are completed with simple ELF descriptors
from the VGG, AlexNet and Xception networks. These new hybrids are then
compared to their respective ELF variant (Right). Results show that these
simpler gradients can detect systematic keypoints with comparable \textit{rep}
on very structured images such as HPatches. However, the ELF detector better
overcomes light changes (Webcam). On HPatches, the Laplacian-variant reaches
similar \textit{ms} as ELF-VGG (55 \textit{vs} 56) and outperforms ELF-AlexNet
and ELF-Xception. These scores can be explained with the images structure: for
heavy textured images, high intensity gradient locations are relevant enough
keypoints. However, on Webcam, all ELF detectors outperform Laplacian and
Sobel with a factor of 100\%. This shows that ELF is more robust than
Laplacian and Sobel operators. Also, feature gradient is a sparse signal which
is better suited for local maxima detection than the much smoother Laplacian
operator (Figure \ref{fig:sobel_visu}).

\textbf{Qualitative results}
Green lines show putative matches based only on nearest neighbour
matching of descriptors.
More qualitative results are available in the video
\footnote{\url{https://youtu.be/oxbG5162yDs}}.
\begin{figure}[thb]
 \centering
 \includegraphics[width=\linewidth]{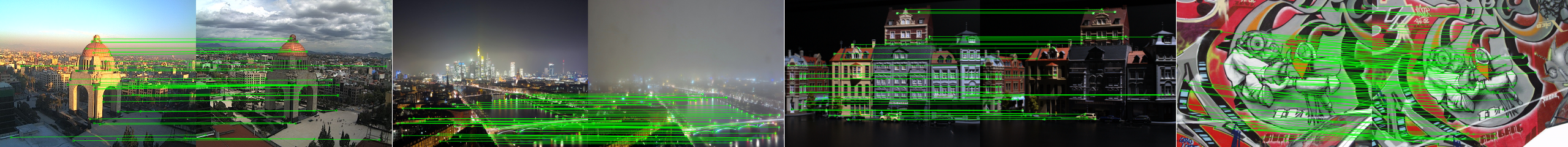}
 \caption{Green lines show putative matches of the
 simple descriptor before RANSAC-based homography estimation.} 
 \label{fig:matching_pic}
\end{figure}

\textbf{CVPR19 Image Matching Challenge \cite{cvpr19challenge}}
This challenge evaluates detection/description methods on two standard tasks:
1) wide stereo matching and 2) structure from motion from small image sets. The
\textit{matching score} evaluates the first task, and the camera pose estimation
is used for both tasks. Both applications are evaluated on the photo-tourism
image collections of popular landmarks \cite{thomee59yfcc100m,
heinly2015reconstructing}. More details on the metrics definition are available
on the challenge website \cite{cvpr19challenge}.

\textit{Wide stereo matching:} Task 1 matches image pairs across wide
baselines.
It is evaluated with the keypoints \textit{ms} and the
relative camera pose estimation between two images. The evaluators
run COLMAP to reconstruct dense `ground-truth' depth which they use to
translate keypoints from one image to another and compute the matching score.
They use the RANSAC inliers to estimate the camera pose and measure performance
with the ``angular difference between the estimated and ground-truth vectors for
both rotation and translation. To reduce this to one value, they use a variable
threshold to determine each pose as correct or not, then compute the area under
the curve up to the angular threshold. This value is thus the mean average
precision up to x, or mAPx. They consider 5, 10, 15, 20, and 25 degrees"
\cite{cvpr19challenge}.
Submissions can contain up to 8000 keypoints and we submitted entries to the
sparse category i.e. methods with up to 512 keypoints.

\begin{figure}[thb]
 \centering
 \includegraphics[width=\linewidth]{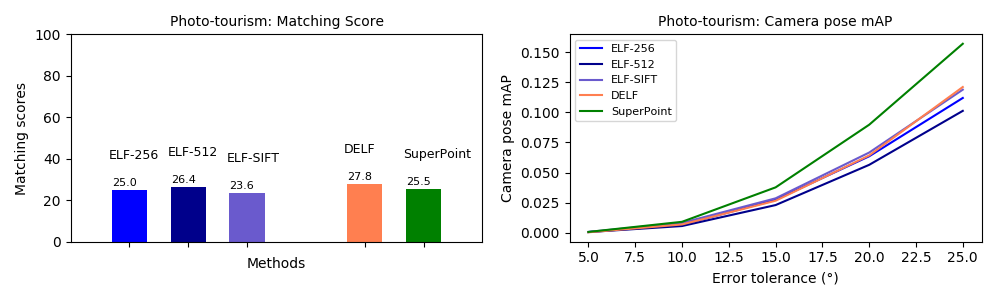}
  \caption{\textit{Wide stereo matching.} Left: matching score (\%) of sparse
  methods (up to 512 keypoints) on photo-tourism. Right: Evolution of mAP of
  camera pose for increasing tolerance threshold (degrees).}
 \label{fig:cvpr19_task1}
\end{figure}

Figure \ref{fig:cvpr19_task1} (left) shows the \textit{ms} (\%) of the
submitted sparse methods. It compares ELF-VGG
detection with DELF \cite{noh2017largescale} and SuperPoint, where ELF is
completed with either the simple descriptor from pool3 or pool4, and SIFT. The
variant are dubbed respectively ELF-256, ELF-512 and ELF-SIFT. This allows us
to sketch a simple comparison of descriptor performances between the
simple descriptor and standard SIFT. 

As previously observed on HPatches and Webcam, ELF and SuperPoint reach similar
scores on Photo-Tourism. ELF-performance slightly increases from 25\% to
26.4\% when switching descriptors from VGG-pool3 to VGG-pool4. One explanation
is that the feature space size is doubled from the first to the second. This
would allow the pool4 descriptors to be more discriminative. However, the 1.4\%
gain may not be worth the additional memory use. Overall, the results show that
ELF can compare with the SoA on this additional dataset that exhibits more
illumination and viewpoint changes than HPatches and Webcam.

This observation is reinforced by the camera pose evaluation (Figure
\ref{fig:cvpr19_task1} right). SuperPoint shows as slight advantage over others
that increases from 1\% to 5\% across the error tolerance threshold
whereas ELF-256 exhibits a minor under-performance. Still, these results show
ELF compares with SoA performance even though it is not trained
explicitly for detection/description.

\begin{figure}[thb]
 \centering
 \includegraphics[width=0.7\linewidth]{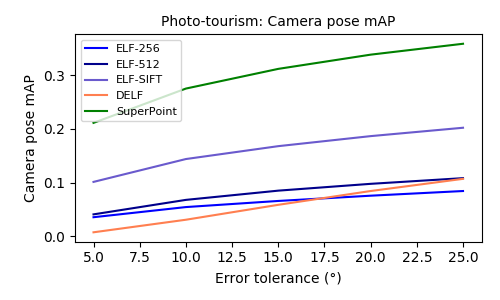}
  \caption{\textit{SfM from small subsets}. Evolution of mAP of camera pose
  for increasing tolerance threshold.}
 \label{fig:cvpr19_task2}
\end{figure}

\textit{Structure-from-Motion from small subsets.}
Task 2 ``proposes to to build SfM reconstructions from small (3, 5, 10, 25)
subsets of images and use the poses obtained from the entire (much larger) set
as ground truth" \cite{cvpr19challenge}.

Figure \ref{fig:cvpr19_task2} shows that SuperPoint reaches performance twice
as big as the next best method ELF-SIFT. This suggests that when few images are
available, SuperPoint performs better than other approaches. One explanation is
that even in 'sparse-mode', \textit{i.e.} when the number of keypoints is restricted up
to 512, SuperPoint samples points more densely than the others ($\sim$383
\textit{v.s.} $\sim$210 for the others). Thus, SuperPoint provides more
keypoints to triangulate i.e. more 2D-3D correspondences to use when
estimating the camera pose. This suggests that high keypoint density is a
crucial characteristic of the detection method for Structure-from-Motion. In
this regard, ELF still has room for improvement compared to SuperPoint. 

\section{Conclusion}
We have introduced ELF, a novel method to extract feature locations
from pre-trained CNNs, with no further training. Extensive experiments show
that it performs as well as state-of-the art detectors. It can easily be
integrated into existing matching pipelines and proves to boost their
matching performances. Even when completed with a simple feature-map-based
descriptor, it turns into a competitive feature matching pipeline. These results
shed new light on the information embedded inside trained CNNs. This work also
raises questions on the descriptor training of deep-learning approaches:
whether their losses actually constrain the CNN to learn better features than
the ones it would learn on its own to complete a vision task. Preliminary
results show that the CNN architecture, the training task and the dataset have
substantial impact on the detector performances. A further analysis of these
correlations is the object of a future work. 

{\small
\bibliographystyle{acm}
\bibliography{../egbib}

\begin{thebibliography}{10}

\bibitem{cvpr19challenge}
Cvpr19 image matching challenge.
\newblock \url{https://image-matching-workshop.github.io/challenge/}, 2019.

\bibitem{aanaes2012interesting}
{\sc Aan{\ae}s, H., Dahl, A.~L., and Pedersen, K.~S.}
\newblock Interesting interest points.
\newblock {\em International Journal of Computer Vision 97}, 1 (2012), 18--35.

\bibitem{abadi2016tensorflow}
{\sc Abadi, M., Barham, P., Chen, J., Chen, Z., Davis, A., Dean, J., Devin, M.,
  Ghemawat, S., Irving, G., Isard, M., et~al.}
\newblock Tensorflow: a system for large-scale machine learning.
\newblock In {\em OSDI\/} (2016), vol.~16, pp.~265--283.

\bibitem{alcantarilla2012kaze}
{\sc Alcantarilla, P.~F., Bartoli, A., and Davison, A.~J.}
\newblock Kaze features.
\newblock In {\em European Conference on Computer Vision\/} (2012), Springer,
  pp.~214--227.

\bibitem{balntas2017hpatches}
{\sc Balntas, V., Lenc, K., Vedaldi, A., and Mikolajczyk, K.}
\newblock Hpatches: A benchmark and evaluation of handcrafted and learned local
  descriptors.
\newblock In {\em Conference on Computer Vision and Pattern Recognition
  (CVPR)\/} (2017), vol.~4, p.~6.

\bibitem{balntas2016learning}
{\sc Balntas, V., Riba, E., Ponsa, D., and Mikolajczyk, K.}
\newblock Learning local feature descriptors with triplets and
  shallowconvolutional neural networks.
\newblock In {\em BMVC\/} (2016), vol.~1, p.~3.

\bibitem{bay2006surf}
{\sc Bay, H., Tuytelaars, T., and Van~Gool, L.}
\newblock Surf: Speeded up robust features.
\newblock In {\em European conference on computer vision\/} (2006), Springer,
  pp.~404--417.

\bibitem{calonder2010brief}
{\sc Calonder, M., Lepetit, V., Strecha, C., and Fua, P.}
\newblock Brief: Binary robust independent elementary features.
\newblock In {\em European conference on computer vision\/} (2010), Springer,
  pp.~778--792.

\bibitem{chollet17xception}
{\sc Chollet, F.}
\newblock Xception: Deep learning with depthwise separable convolutions.
\newblock In {\em 2017 {IEEE} Conference on Computer Vision and Pattern
  Recognition, {CVPR} 2017, Honolulu, HI, USA, July 21-26, 2017\/} (2017),
  pp.~1800--1807.

\bibitem{choy2016universal}
{\sc Choy, C.~B., Gwak, J., Savarese, S., and Chandraker, M.}
\newblock Universal correspondence network.
\newblock In {\em Advances in Neural Information Processing Systems\/} (2016),
  pp.~2414--2422.

\bibitem{deng2009imagenet}
{\sc Deng, J., Dong, W., Socher, R., Li, L.-J., Li, K., and Fei-Fei, L.}
\newblock Imagenet: A large-scale hierarchical image database.
\newblock In {\em Computer Vision and Pattern Recognition, 2009. CVPR 2009.
  IEEE Conference on\/} (2009), Ieee, pp.~248--255.

\bibitem{detone18superpoint}
{\sc DeTone, D., Malisiewicz, T., and Rabinovich, A.}
\newblock Superpoint: Self-supervised interest point detection and description.
\newblock In {\em CVPR Deep Learning for Visual SLAM Workshop\/} (2018).

\bibitem{fischer2014descriptor}
{\sc Fischer, P., Dosovitskiy, A., and Brox, T.}
\newblock Descriptor matching with convolutional neural networks: a comparison
  to sift.
\newblock {\em arXiv preprint arXiv:1405.5769\/} (2014).

\bibitem{gatys2016image}
{\sc Gatys, L.~A., Ecker, A.~S., and Bethge, M.}
\newblock Image style transfer using convolutional neural networks.
\newblock In {\em Proceedings of the IEEE Conference on Computer Vision and
  Pattern Recognition\/} (2016), pp.~2414--2423.

\bibitem{han2015matchnet}
{\sc Han, X., Leung, T., Jia, Y., Sukthankar, R., and Berg, A.~C.}
\newblock Matchnet: Unifying feature and metric learning for patch-based
  matching.
\newblock In {\em Proceedings of the IEEE Conference on Computer Vision and
  Pattern Recognition\/} (2015), pp.~3279--3286.

\bibitem{heinly2015reconstructing}
{\sc Heinly, J., Schonberger, J.~L., Dunn, E., and Frahm, J.-M.}
\newblock Reconstructing the world* in six days*(as captured by the yahoo 100
  million image dataset).
\newblock In {\em Proceedings of the IEEE Conference on Computer Vision and
  Pattern Recognition\/} (2015), pp.~3287--3295.

\bibitem{kapur1985new}
{\sc Kapur, J.~N., Sahoo, P.~K., and Wong, A.~K.}
\newblock A new method for gray-level picture thresholding using the entropy of
  the histogram.
\newblock {\em Computer vision, graphics, and image processing 29}, 3 (1985),
  273--285.

\bibitem{krizhevsky2012imagenet}
{\sc Krizhevsky, A., Sutskever, I., and Hinton, G.~E.}
\newblock Imagenet classification with deep convolutional neural networks.
\newblock In {\em Advances in neural information processing systems\/} (2012),
  pp.~1097--1105.

\bibitem{lenc12vlbenchmarks}
{\sc Lenc, K., Gulshan, V., and Vedaldi, A.}
\newblock Vlbenchmkars.
\newblock \url{http://www.vlfeat.org/benchmarks/}xsxs, 2011.

\bibitem{lin2014microsoft}
{\sc Lin, T.-Y., Maire, M., Belongie, S., Hays, J., Perona, P., Ramanan, D.,
  Doll{\'a}r, P., and Zitnick, C.~L.}
\newblock Microsoft coco: Common objects in context.
\newblock In {\em European conference on computer vision\/} (2014), Springer,
  pp.~740--755.

\bibitem{lowe2004distinctive}
{\sc Lowe, D.~G.}
\newblock Distinctive image features from scale-invariant keypoints.
\newblock {\em International journal of computer vision 60}, 2 (2004), 91--110.

\bibitem{mahendran2015understanding}
{\sc Mahendran, A., and Vedaldi, A.}
\newblock Understanding deep image representations by inverting them.
\newblock In {\em Proceedings of the IEEE conference on computer vision and
  pattern recognition\/} (2015), pp.~5188--5196.

\bibitem{matas2004robust}
{\sc Matas, J., Chum, O., Urban, M., and Pajdla, T.}
\newblock Robust wide-baseline stereo from maximally stable extremal regions.
\newblock {\em Image and vision computing 22}, 10 (2004), 761--767.

\bibitem{melekhov2016siamese}
{\sc Melekhov, I., Kannala, J., and Rahtu, E.}
\newblock Siamese network features for image matching.
\newblock In {\em 2016 23rd International Conference on Pattern Recognition
  (ICPR)\/} (2016), IEEE, pp.~378--383.

\bibitem{mikolajczyk2005performance}
{\sc Mikolajczyk, K., and Schmid, C.}
\newblock A performance evaluation of local descriptors.
\newblock {\em IEEE transactions on pattern analysis and machine intelligence
  27}, 10 (2005), 1615--1630.

\bibitem{mikolajczyk2005comparison}
{\sc Mikolajczyk, K., Tuytelaars, T., Schmid, C., Zisserman, A., Matas, J.,
  Schaffalitzky, F., Kadir, T., and Van~Gool, L.}
\newblock A comparison of affine region detectors.
\newblock {\em International journal of computer vision 65}, 1-2 (2005),
  43--72.

\bibitem{noh2017largescale}
{\sc Noh, H., Araujo, A., Sim, J., Weyand, T., and Han, B.}
\newblock Largescale image retrieval with attentive deep local features.
\newblock In {\em Proceedings of the IEEE International Conference on Computer
  Vision\/} (2017), pp.~3456--3465.

\bibitem{ono2018lf}
{\sc Ono, Y., Trulls, E., Fua, P., and K.M.Yi}.
\newblock Lf-net: Learning local features from images.
\newblock In {\em Advances in Neural Information Processing Systems\/} (2018).

\bibitem{rosten2006machine}
{\sc Rosten, E., and Drummond, T.}
\newblock Machine learning for high-speed corner detection.
\newblock In {\em European conference on computer vision\/} (2006), Springer,
  pp.~430--443.

\bibitem{rublee2011orb}
{\sc Rublee, E., Rabaud, V., Konolige, K., and Bradski, G.}
\newblock Orb: An efficient alternative to sift or surf.
\newblock In {\em Computer Vision (ICCV), 2011 IEEE international conference
  on\/} (2011), IEEE, pp.~2564--2571.

\bibitem{savinov2017quad}
{\sc Savinov, N., Seki, A., Ladicky, L., Sattler, T., and Pollefeys, M.}
\newblock Quad-networks: unsupervised learning to rank for interest point
  detection.
\newblock In {\em Proc. IEEE Conference on Computer Vision and Pattern
  Recognition (CVPR)\/} (2017).

\bibitem{schonberger2016structure}
{\sc Schonberger, J.~L., and Frahm, J.-M.}
\newblock Structure-from-motion revisited.
\newblock In {\em Proceedings of the IEEE Conference on Computer Vision and
  Pattern Recognition\/} (2016), pp.~4104--4113.

\bibitem{selvaraju2017grad}
{\sc Selvaraju, R.~R., Cogswell, M., Das, A., Vedantam, R., Parikh, D., Batra,
  D., et~al.}
\newblock Grad-cam: Visual explanations from deep networks via gradient-based
  localization.
\newblock In {\em ICCV\/} (2017), pp.~618--626.

\bibitem{simo2015discriminative}
{\sc Simo-Serra, E., Trulls, E., Ferraz, L., Kokkinos, I., Fua, P., and
  Moreno-Noguer, F.}
\newblock Discriminative learning of deep convolutional feature point
  descriptors.
\newblock In {\em Proceedings of the IEEE International Conference on Computer
  Vision\/} (2015), pp.~118--126.

\bibitem{simonyan2013deep}
{\sc Simonyan, K., Vedaldi, A., and Zisserman, A.}
\newblock Deep inside convolutional networks: Visualising image classification
  models and saliency maps.
\newblock {\em arXiv preprint arXiv:1312.6034\/} (2013).

\bibitem{simonyan2014very}
{\sc Simonyan, K., and Zisserman, A.}
\newblock Very deep convolutional networks for large-scale image recognition.
\newblock {\em arXiv preprint arXiv:1409.1556\/} (2014).

\bibitem{smilkov2017smoothgrad}
{\sc Smilkov, D., Thorat, N., Kim, B., Vi{\'e}gas, F., and Wattenberg, M.}
\newblock Smoothgrad: removing noise by adding noise.
\newblock {\em arXiv preprint arXiv:1706.03825\/} (2017).

\bibitem{springenberg2015striving}
{\sc Springenberg, J., Dosovitskiy, A., Brox, T., and Riedmiller, M.}
\newblock Striving for simplicity: The all convolutional net.
\newblock In {\em ICLR (workshop track)\/} (2015).

\bibitem{strecha2008benchmarking}
{\sc Strecha, C., Von~Hansen, W., Van~Gool, L., Fua, P., and Thoennessen, U.}
\newblock On benchmarking camera calibration and multi-view stereo for high
  resolution imagery.
\newblock In {\em Computer Vision and Pattern Recognition, 2008. CVPR 2008.
  IEEE Conference on\/} (2008), Ieee, pp.~1--8.

\bibitem{sundararajan2017axiomatic}
{\sc Sundararajan, M., Taly, A., and Yan, Q.}
\newblock Axiomatic attribution for deep networks.
\newblock In {\em International Conference on Machine Learning\/} (2017),
  pp.~3319--3328.

\bibitem{taira2018inloc}
{\sc Taira, H., Okutomi, M., Sattler, T., Cimpoi, M., Pollefeys, M., Sivic, J.,
  Pajdla, T., and Torii, A.}
\newblock Inloc: Indoor visual localization with dense matching and view
  synthesis.
\newblock In {\em Proceedings of the IEEE Conference on Computer Vision and
  Pattern Recognition\/} (2018), pp.~7199--7209.

\bibitem{thomee59yfcc100m}
{\sc Thomee, B., Shamma, D.~A., Friedland, G., Elizalde, B., Ni, K., Poland,
  D., Borth, D., and Li, L.-J.}
\newblock Yfcc100m: The new data in multimedia research.
\newblock {\em Communications of the ACM 59}, 2, 64--73.

\bibitem{verdie2015tilde}
{\sc Verdie, Y., Yi, K., Fua, P., and Lepetit, V.}
\newblock Tilde: A temporally invariant learned detector.
\newblock In {\em Proceedings of the IEEE Conference on Computer Vision and
  Pattern Recognition\/} (2015), pp.~5279--5288.

\bibitem{yi2016lift}
{\sc Yi, K.~M., Trulls, E., Lepetit, V., and Fua, P.}
\newblock Lift: Learned invariant feature transform.
\newblock In {\em European Conference on Computer Vision\/} (2016), Springer,
  pp.~467--483.

\bibitem{zagoruyko2015learning}
{\sc Zagoruyko, S., and Komodakis, N.}
\newblock Learning to compare image patches via convolutional neural networks.
\newblock In {\em Proceedings of the IEEE conference on computer vision and
  pattern recognition\/} (2015), pp.~4353--4361.

\bibitem{zeiler2014visualizing}
{\sc Zeiler, M.~D., and Fergus, R.}
\newblock Visualizing and understanding convolutional networks.
\newblock In {\em European conference on computer vision\/} (2014), Springer,
  pp.~818--833.

\end{thebibliography}
}

\newpage
\appendix

\begin{figure*}[thb]
 \centering
 \includegraphics[width=\linewidth]{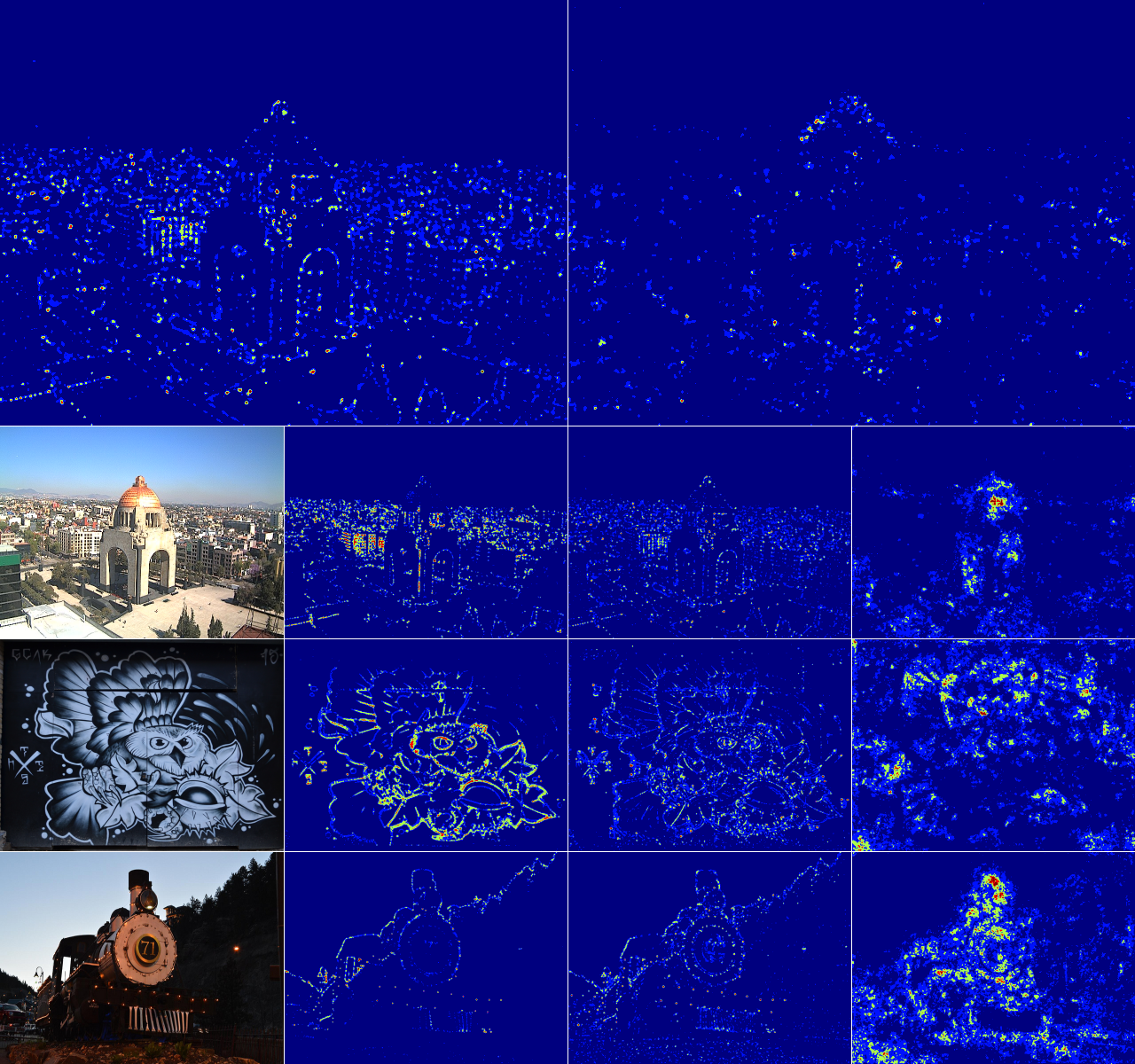}
  \caption{Enlargement of Figure \ref{fig:saliency_coco}.
  Saliency maps computed from the feature map gradient 
 $\left| ^TF^l(x) \cdot \frac{\partial F^l}{\partial \mathbf{I}} \right|$.
 Enhanced image contrast for better visualisation.
 Top row: gradients of VGG $pool_2$ and $pool_3$ show a loss of resolution 
 from $pool_2$ to $pool_3$.
 Bottom: $(pool_i)_{i \in [1,2,5]}$ of VGG on Webcam, HPatches and Coco images.
 Low level saliency maps activate accurately whereas higher saliency maps are blurred.}
 \label{fig:big_saliency_coco}
\end{figure*}

\section{Metrics definition}
We explicit the repeatability and matching score definitions 
introduced in \cite{mikolajczyk2005comparison} and our adaptations 
using the following notations:
let $(\mathbf{I}^1, \mathbf{I}^2)$, be a pair of images and 
$\mathcal{KP}^i = (kp_j^i)_{j<N_i}$ the set of $N_i$ keypoints in image
$\mathbf{I_i}$. Both metrics are in the range $[0,1]$ but we express them as 
percentages for better expressibility.

\paragraph{Repeatability}
Repeatability measures the percentage of keypoints common to both images.
We first warp $\mathcal{KP}^1$ to $\mathbf{I}^2$ and note $\mathcal{KP}^{1,w}$ 
the result.
A naive definition of repeatability is to count the number of pairs 
$(kp^{1,w}, kp^2) \in \mathcal{KP}^{1,w} \times \mathcal{KP}^2$ such that 
$\|kp^{1,w}-kp^2\|_2 < \epsilon$, with $\epsilon$ a distance threshold.
As pointed by \cite{verdie2015tilde}, this
definition overestimates the detection performance for two reasons: a keypoint 
close to several projections can be counted several times. Moreover, with a 
large enough number of keypoints, even simple random sampling can achieve high 
repeatability as the density of the keypoints becomes high.

We instead use the definition implemented in VLBench~\cite{lenc12vlbenchmarks}: 
we define a weighted graph $(V,E)$ where the edges are all 
the possible keypoint pairs between $\mathcal{KP}^{1,w}$ and $\mathcal{KP}^2$ and the 
weights are the euclidean distance between keypoints. 
\begin{equation} \label{eq: graph_dfn}
  \begin{split}
    V &= (kp^{1,w} \in \mathcal{KP}^{1,w}) \cup (kp^2 \in \mathcal{KP}^2) \\
    E &= (kp^{1,w}, kp^2, \|kp^{1,w} - kp^2\|_2) \in \mathcal{KP}^{1,w} \times \mathcal{KP}^2 
  \end{split}
\end{equation}

We run a greedy bipartite matching on the graph and count the matches with a distance 
less than $\epsilon_{kp}$. With $\mathcal{M}$ be the resulting set of matches:

\begin{equation} \label{rep_dfn}
  repeatability = \frac{\mathcal{M}}{\textrm{min}(|\mathcal{KP}^1|, |\mathcal{KP}^2|)}
\end{equation}

We set the distance threshold $\epsilon=5$ as is done in LIFT
\cite{yi2016lift} and LF-Net \cite{ono2018lf}.

\paragraph{Matching score}
The matching score definition introduced in \cite{mikolajczyk2005comparison} 
captures the percentage of keypoint pairs that are nearest neighbours both
in image space and in descriptor space, and for which these two distances are below
their respective threshold $\epsilon_{kp}$ and $\epsilon_{d}$.
Let $\mathcal{M}$ be defined as in the previous paragraph and $\mathcal{M}_d$
be the analog of $\mathcal{M}$ when the graph weights are descriptor distances instead
of keypoint euclidean distances. We delete all the pairs with a distance
above the thresholds $\epsilon$ and $\epsilon_d$ in $\mathcal{M}$ and
$\mathcal{M}_d$ respectively. We then count the number of pairs which are both
nearest neigbours in image space and descriptor space i.e. the intersection of
$\mathcal{M}$ and $\mathcal{M}_d$: 

\begin{equation} \label{MS}
  matching \; score = \frac{\mathcal{M} \cap \mathcal{M}_d}{\textrm{min}(|\mathcal{KP}^1|, |\mathcal{KP}^2|)}
\end{equation}

One drawback of this definition is that there is no unique descriptor distance
threshold $\epsilon_d$ valid for all methods. For example, the SIFT descriptor
as computed by OpenCV is a $[0,255]^{128}$ vector for better computational 
precision, the SuperPoint descriptor is a $[0,1]^{256}$ vector and the ORB
descriptor is a 32 bytes binary vector. Not only the vectors are not defined
over the same normed space but their range vary significantly. To avoid introducing
human bias by setting a descriptor distance threshold $\epsilon_d$ for each
method, we choose to set $\epsilon_d = \infty$ and compute the matching score
as in \cite{mikolajczyk2005comparison}. This means that we consider any
descriptor match valid as long as they match corresponding keypoints even when
the descriptor distance is high.

\section{Tabular results}

\begin{table}[tbh]
  \small
\begin{center}
\begin{tabular}{|l|cc|cc|cc|}
\hline
        & \multicolumn{2}{c|}{Repeatability} & \multicolumn{2}{c|}{Matching Score} \\
        & \cite{balntas2017hpatches} & \cite{verdie2015tilde}& 
        \cite{balntas2017hpatches} & \cite{verdie2015tilde} \\
\hline
  ELF-VGG       & 63.81  & \textcolor{mygreen}{\textbf{53.23}} & 
  \textcolor{blue}{\textbf{51.84}} & \textcolor{mygreen}{\textbf{43.73}} \\
  ELF-AlexNet   & 51.30           & 38.54          & 35.21          & 31.92 \\
  ELF-Xception  & 48.06           & \textcolor{blue}{\textbf{49.84}} & 29.81 & \textcolor{blue}{\textbf{35.48}} \\

  ELF-SuperPoint  & 59.7         & 46.29 & 44.32  & 18.11  \\
  ELF-LFNet       & 60.1         & 41.90 & 44.56  & 33.43 \\
\hline
  LF-Net    & 61.16           & 48.27          & 34.19          & 18.10 \\
  SuperPoint& \textcolor{mygreen}{\textbf{68.57}}  & 46.35  & \textcolor{mygreen}{\textbf{57.11}} & 32.44 \\
  LIFT      & 54.66           & 42.21          & 34.02          & 17.83 \\
  SURF      & 54.51           & 33.93          & 26.10          & 10.13 \\
  SIFT      & 51.19           & 28.25          & 24.58          & 8.30 \\
  ORB       & 53.44           & 31.56          & 14.76          & 1.28 \\
  KAZE      & 56.88           & 41.04          & 29.81          & 13.88 \\
  TILDE     & \textcolor{blue}{\textbf{65.96}}
  & 52.53          & 46.71          & 34.67 \\
  MSER      & 47.82           & 52.23          & 21.08          & 6.14 \\
\hline
\end{tabular}
\end{center}
  \caption{Generic performances on HPatches \cite{balntas2017hpatches}. 
  Robustness to light (Webcam \cite{verdie2015tilde}). (Fig. 5).}
\label{tab:whole_pipeline}
\end{table}

\begin{table}[tbh]
  \small
  \begin{center}
    \begin{tabular}{|c|c|c|c|c|c|c|}
      \hline
      & \scriptsize{LF-Net} & \scriptsize{SuperPoint} & \small{LIFT} & \small{SIFT}  & \small{SURF}  & \small{ORB}  \\
      \hline
      \multirow{2}{*}{\scriptsize{\cite{balntas2017hpatches}}} 
      & 34.19          & \textbf{57.11} & 34.02         & 24.58         & 26.10         & 14.76 \\
      & \textbf{44.19} & 53.71          & \textbf{39.48}& \textbf{27.03}& \textbf{34.97}& \textbf{20.04}  \\
      \hline
      \multirow{2}{*}{\scriptsize{\cite{verdie2015tilde}}}
      & 18.10 & 32.44         & 17.83          & 10.13         & 8.30          & 1.28 \\
      & \textbf{30.71} & \textbf{34.60}& \textbf{26.84} & \textbf{13.21}& \textbf{21.43} & \textbf{13.91}  \\
      \hline
    \end{tabular}
  \end{center}
  \caption{Individual component performance
  (Fig.~\ref{fig:ind_component}-stripes). Matching score for the integration of
  the VGG $pool_3$ simple-descriptor with other's detection. Top: Original
  description. Bottom: Integration of simple-descriptor. HPatches:
  \cite{balntas2017hpatches}. Webcam: \cite{verdie2015tilde}}
  \label{tab:cross_res_des}
\end{table}

\begin{table}[tbh]
  \small
  \begin{center}
    \begin{tabular}{|c|c|c|c|c|c|c|}
      \hline
      & \scriptsize{LF-Net} & \scriptsize{SuperPoint} & \small{LIFT} & \small{SIFT}  & \small{SURF}  & \small{ORB}  \\
      \hline
      \multirow{2}{*}{\scriptsize{\cite{balntas2017hpatches}}} 
      & 34.19          & \textbf{57.11} & 34.02          & 24.58          & 26.10         & 14.76 \\
      & \textbf{39.16} & 54.44          & \textbf{42.48} & \textbf{50.63} & \textbf{30.91}& \textbf{36.96}\\
      \hline
      \multirow{2}{*}{\scriptsize{\cite{verdie2015tilde}}}
      & 18.10          & 32.44         & 17.83          & 10.13          & 8.30           & 1.28 \\
      &\textbf{26.70} & \textbf{39.55}& \textbf{30.82} & \textbf{36.83} & \textbf{19.14} & \textbf{6.60} \\
      \hline
    \end{tabular}
  \end{center}
  \caption{Individual component performance
  (Fig.~\ref{fig:ind_component}-circle). Matching score for the integration of
  ELF-VGG (on $pool_2$) with other's descriptor. Top: Original detection.
  Bottom: Integration of ELF. HPatches: \cite{balntas2017hpatches}. Webcam:
  \cite{verdie2015tilde}}
\label{tab:cross_res_det}
\end{table}

\begin{table}[tbh]
\begin{center}
\begin{tabular}{|l|cc|cc|cc|}
\hline
        & \multicolumn{2}{c|}{Repeatability} & \multicolumn{2}{c|}{Matching Score} \\
        & \cite{balntas2017hpatches} & \cite{verdie2015tilde}& 
        \cite{balntas2017hpatches} & \cite{verdie2015tilde} \\
\hline
  Sobel-VGG     & 56.99           & 33.74          & 42.11          & 20.99  \\
  Lapl.-VGG     & \textbf{65.45}  & 33.74          & \textbf{55.25} & 22.79  \\
  VGG           & 63.81           & \textbf{53.23} & 51.84          & \textbf{43.73} \\
  \hline
  Sobel-AlexNet & 56.44           & 33.74          & 30.57          & 15.42 \\
  Lapl.-AlexNet & \textbf{65.93}  & 33.74          & \textbf{40.92} & 15.42  \\
  AlexNet       & 51.30           & \textbf{38.54} & 35.21          & \textbf{31.92} \\
  \hline
  Sobel-Xception& 56.44           & 33.74          & 34.14          & 16.86 \\
  Lapl.-Xception& \textbf{65.93}  & 33.74          & \textbf{42.52} & 16.86  \\
  Xception      & 48.06           & \textbf{49.84} & 29.81          & \textbf{35.48} \\
\hline
\end{tabular}
\end{center}
  \caption{Gradient baseline on HPatches \cite{balntas2017hpatches}
  and Webcam \cite{verdie2015tilde} (Fig. \ref{fig:gradient_perf} ).}
\label{tab:cmp_sobel}
\end{table}

\section{ELF Meta Parameters}
This section specifies the meta parameters values for the ELF variants.
For all methods, $(w_{NMS}, b_{NMS})=(10,10)$.

\begin{itemize}
  \item Denoise: $(\mu_{noise}, \sigma_{noise})$.
  \item Threshold: $(\mu_{thr}, \sigma_{thr})$.
  \item $F^l$: the feature map which gradient is used for detection.
  \item simple-des: the feature map used for simple-description. Unless mentioned
    otherwise, the feature map is taken from the same network as the detection
    feature map $F^l$.
\end{itemize}

\begin{table}[tbh]
 \scriptsize
 \begin{center}
   \begin{tabular}{|l|c|c|c|c|c|}
     \hline
     Nets       & Denoise   & Threshold & $F^l$         & simple-desc \\
     \hline
     VGG        & (5,5)     & (5,4)     & pool2         & pool4     \\
     \hline
     Alexnet    & (5,5)     & (5,4)    & pool1         & pool2   \\
     \hline
     Xception   & (9,3)     & (5,4)     & block2-conv1  & block4-pool \\
     \hline
     SuperPoint & (7,2)     & (17,6)    & conv1a        & VGG-pool3 \\
     \hline
     LF-Net     & (5,5)     & (5,4)     & block2-BN     & VGG-pool3 \\
     \hline
   \end{tabular}
 \end{center}
  \caption{Generic performances on HPatches (Fig. \ref{fig:hpatch_gle_perf}). (BN: Batch Norm)}
  \label{tab:meta_params}
\end{table}

\begin{table}[tbh]
 \scriptsize
 \begin{center}
   \begin{tabular}{|l|c|c|c|c|c|}
     \hline
     Nets       & Denoise   & Threshold & $F^l$         & simple-desc \\
     \hline
     VGG        & (5,5)     & (5,4)     & pool2         & pool4     \\
     \hline
     Alexnet    & (5,5)     & (5,4)     & pool1         & pool2   \\
     \hline
     Xception   & (9,9)     & (5,4)     & block2-conv1  & block4-pool \\
     \hline
     SuperPoint & (7,2)     & (17,6)    & conv1a        & VGG-pool3 \\
     \hline
     LF-Net     & (5,5)     & (5,4)     & block2-conv   & VGG-pool3 \\
     \hline
   \end{tabular}
 \end{center}
  \caption{Robustness to light on Webcam (Fig. \ref{fig:hpatch_gle_perf}).}
  \label{tab:meta_params}
\end{table}

\begin{table}[tbh]
 \scriptsize
 \begin{center}
   \begin{tabular}{|l|c|c|c|c|c|}
     \hline
     Nets       & Denoise   & Threshold & $F^l$   & simple-desc \\
     \hline
     VGG        & (5,2)     & (17,6)    & pool2   & pool4     \\
     \hline
   \end{tabular}
 \end{center}
  \caption{Robustness to scale on HPatches (Fig. \ref{fig:robust_scale}).}
  \label{tab:meta_params}
\end{table}

\begin{table}[tbh]
 \scriptsize
 \begin{center}
   \begin{tabular}{|l|c|c|c|c|c|}
     \hline
     Nets       & Denoise   & Threshold & $F^l$   & simple-desc \\
     \hline
     VGG        & (5,2)     & (17,6)    & pool2   & pool4     \\
     \hline
   \end{tabular}
 \end{center}
  \caption{Robustness to rotation on HPatches (Fig. \ref{fig:robust_rotation}).}
  \label{tab:meta_params}
\end{table}

\begin{table}[tbh]
 \scriptsize
 \begin{center}
   \begin{tabular}{|l|c|c|c|c|c|}
     \hline
     Nets       & Denoise   & Threshold & $F^l$   & simple-desc \\
     \hline
     VGG        & (5,2)     & (17,6)    & pool2   & pool4     \\
     \hline
   \end{tabular}
 \end{center}
  \caption{Robustness to 3D viewpoint on Strecha (Fig. \ref{fig:robust_strecha}).}
  \label{tab:meta_params}
\end{table}

\begin{table}[tbh]
 \scriptsize
 \begin{center}
   \begin{tabular}{|l|c|c|c|c|c|}
     \hline
     Nets       & Denoise   & Threshold & $F^l$   & simple-desc \\
     \hline
     VGG        & (5,5)     & (5,5)    & pool2   & pool3     \\
     \hline
   \end{tabular}
 \end{center}
  \caption{Individual component analysis (Fig. \ref{fig:ind_component})}
  \label{tab:meta_params}
\end{table}

\begin{table}[tbh]
 \scriptsize
 \begin{center}
   \begin{tabular}{|l|c|c|c|c|c|}
     \hline
     Nets       & Denoise   & Threshold & $F^l$   & simple-desc \\
     \hline
     VGG        & (5,5)     & (5,4)    & pool2   & pool4     \\
     \hline
     Sobel      & (9,9)     & (5,4)    & -   & pool4     \\
     \hline
     Laplacian  & (9,9)     & (5,4)    & -   & pool4     \\
     \hline
   \end{tabular}
 \end{center}
  \caption{Gradient baseline on HPatches and Webcam (Fig.
  \ref{fig:gradient_perf}).}
  \label{tab:meta_params}
\end{table}

\end{document}